\documentclass[sigconf]{acmart}

\usepackage{bm,multirow,pifont}

\newcommand{\cmark}{\ding{51}}
\newcommand{\xmark}{\ding{55}}
\usepackage{color, colortbl}
\definecolor{LightGray}{gray}{0.9}

\newcolumntype{L}[1]{>{\raggedright\let\newline\\\arraybackslash\hspace{0pt}}m{#1}}
\newcolumntype{C}[1]{>{\centering\let\newline\\\arraybackslash\hspace{0pt}}m{#1}}
\newcolumntype{R}[1]{>{\raggedleft\let\newline\\\arraybackslash\hspace{0pt}}m{#1}}

\AtBeginDocument{%
  }

\copyrightyear{2024}
\acmYear{2024}
\setcopyright{acmlicensed}
\acmConference[MM '24]{Proceedings of the 32nd ACM International Conference on Multimedia}{October 28-November 1, 2024}{Melbourne, VIC, Australia}
\acmBooktitle{Proceedings of the 32nd ACM International Conference on Multimedia (MM '24), October 28-November 1, 2024, Melbourne, VIC, Australia}
\acmDOI{10.1145/3664647.3680557}
\acmISBN{979-8-4007-0686-8/24/10}
\begin{document}

\title{iControl3D: An Interactive System for Controllable 3D Scene Generation}

\author{Xingyi Li}
\orcid{0000-0001-5765-3852}
\affiliation{%
  \institution{School of AIA, Huazhong University of Science and Technology}
  \city{Wuhan}
  \country{China}
  }
\affiliation{%
  \institution{S-Lab, Nanyang Technological University}
  \city{Singapore}
  \country{Singapore}
  }
\email{xingyi_li@hust.edu.cn}

\author{Yizheng Wu}
\orcid{0000-0001-5335-4919}
\affiliation{%
  \institution{School of AIA, Huazhong University of Science and Technology}
  \city{Wuhan}
  \country{China}
  }
\affiliation{%
  \institution{S-Lab, Nanyang Technological University}
  \city{Singapore}
  \country{Singapore}
  }
\email{yzwu21@hust.edu.cn}

\author{Jun Cen}
\orcid{0000-0002-7578-7667}
\affiliation{%
  \institution{S-Lab, Nanyang Technological University}
  \city{Singapore}
  \country{Singapore}}
\email{jcenaa@connect.ust.hk}

\author{Juewen Peng}
\orcid{0000-0001-5740-2682}
\affiliation{%
  \institution{College of Computing and Data Science, Nanyang Technological University}
  \city{Singapore}
  \country{Singapore}}
\email{juewen.peng@ntu.edu.sg}

\author{Kewei Wang}
\orcid{0000-0001-9433-720X}
\affiliation{%
  \institution{School of AIA, Huazhong University of Science and Technology}
  \city{Wuhan}
  \country{China}
  }
\affiliation{%
  \institution{S-Lab, Nanyang Technological University}
  \city{Singapore}
  \country{Singapore}
  }
\email{wangkewei@hust.edu.cn}

\author{Ke Xian}
\orcid{0000-0002-0884-5126}
\affiliation{%
  \institution{School of EIC, Huazhong University of Science and Technology}
  \city{Wuhan}
  \country{China}}
\email{kxian@hust.edu.cn}

\author{Zhe Wang}
\orcid{0000-0002-0597-4475}
\affiliation{%
  \institution{SenseTime Research}
  \city{Hong Kong SAR}
  \country{China}}
\email{wangzhe@sensetime.com}

\author{Zhiguo Cao}
\orcid{0000-0002-9223-1863}
\authornote{Corresponding author.}
\affiliation{%
  \institution{School of AIA, Huazhong University of Science and Technology}
  \city{Wuhan}
  \country{China}}
\email{zgcao@hust.edu.cn}

\author{Guosheng Lin}
\orcid{0000-0002-0329-7458}
\affiliation{%
  \institution{S-Lab, Nanyang Technological University}
  \city{Singapore}
  \country{Singapore}}
\email{gslin@ntu.edu.sg}

\renewcommand{\shortauthors}{Xingyi Li et al.}

\begin{abstract}
  3D content creation has long been a complex and time-consuming process, often requiring specialized skills and resources. While recent advancements have allowed for text-guided 3D object and scene generation, they still fall short of providing sufficient control over the generation process, leading to a gap between the user's creative vision and the generated results. In this paper, we present iControl3D, a novel interactive system that empowers users to generate and render customizable 3D scenes with precise control. To this end, a 3D creator interface has been developed to provide users with fine-grained control over the creation process. Technically, we leverage 3D meshes as an intermediary proxy to iteratively merge individual 2D diffusion-generated images into a cohesive and unified 3D scene representation. To ensure seamless integration of 3D meshes, we propose to perform boundary-aware depth alignment before fusing the newly generated mesh with the existing one in 3D space. Additionally, to effectively manage depth discrepancies between remote content and foreground, we propose to model remote content separately with an environment map instead of 3D meshes. Finally, our neural rendering interface enables users to build a radiance field of their scene online and navigate the entire scene. Extensive experiments have been conducted to demonstrate the effectiveness of our system. The code will be made available at \url{https://github.com/xingyi-li/iControl3D}.
\end{abstract}




\maketitle

\begin{figure*}
  \includegraphics[width=\textwidth]{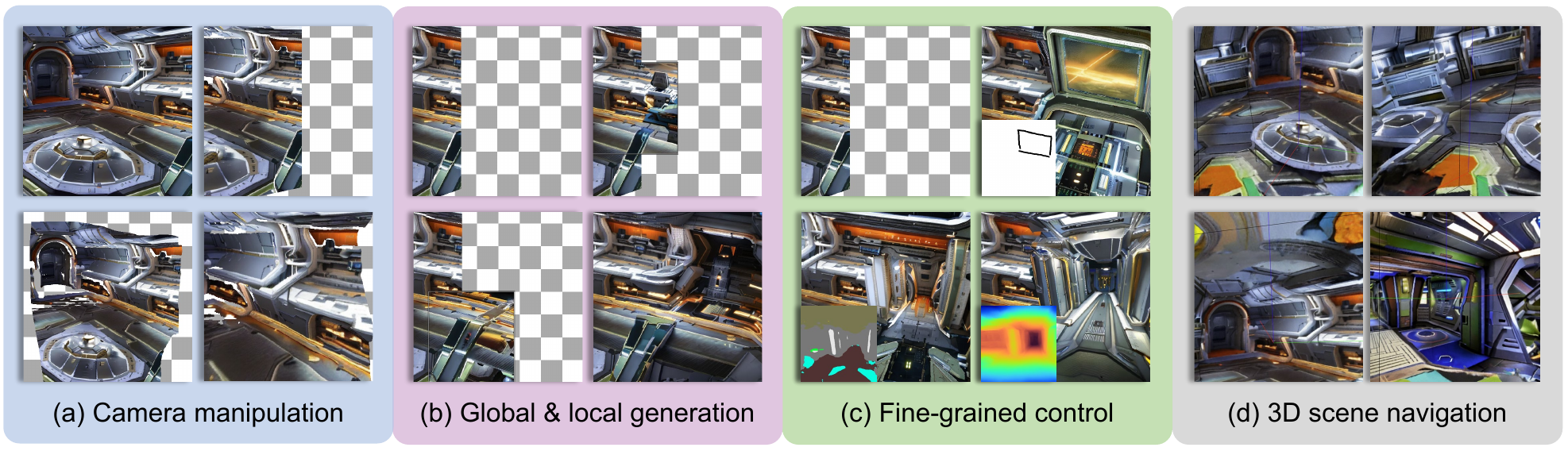}
  \caption{Our system empowers users to generate and render customizable 3D scenes with precise control over the 3D scene generation process. With our system, users can actively participate in the 3D scene creation process. For example, they can (a) manipulate the virtual camera to any viewpoint, (b) adjust the size of the selection box to generate global and local content, and try different random seeds to generate various results. (c) Besides text prompts, users can achieve fine-grained control over the output by adding extra conditions such as scribbles, semantic segmentation maps, and depth. (d) After generating 3D scenes, they can navigate the entire scene and create camera trajectories to render videos according to their preferences.}
  \label{fig:teaser}
\end{figure*}

\section{Introduction}
Recent years have witnessed explosive growth in the development of generative image and video models. In particular, diffusion models~\cite{sohl2015deep,ho2020denoising,Rombach_2022_CVPR,dhariwal2021diffusion} have pushed the boundaries of image generation, or AI-Generated Content (AIGC) to an unprecedented level of realism, with their outputs often indistinguishable from real images. Despite the success in the 2D domain, 
generating 3D assets and realistic 3D scenes remains a complex process that requires a significant amount of expertise and specialized software. It can take years of practice to master the necessary skills and techniques involved in the 3D content creation. 

In light of this, many researchers are eager to extend the power of 2D diffusion models to the field of 3D generation. Existing works~\cite{poole2022dreamfusion,lin2022magic3d,wang2023prolificdreamer,liu2023one,liu2023zero} have demonstrated the potential of text-guided 3D object generation using 2D diffusion. Yet, these methods present challenges when it comes to generating 3D structures and textures on a scene-scale level. 
Inspired by previous studies~\cite{liu2021infinite,li2022infinitenature,cai2022diffdreamer}, Fridman et al.~\cite{fridman2023scenescape} introduce SceneScape, a novel method for text-driven perpetual view generation. 
While SceneScape enables the synthesis of flying-out trajectories of scenes from text, 
it struggles with generating complete 3D scenes. 
Concurrently, Text2Room~\cite{hollein2023text2room} proposes to create room-scale textured 3D meshes by using pre-trained 2D text-to-image diffusion models. However, it is restricted to indoor scene generation and offers limited control over the synthesis process, since only text and predefined camera trajectory are available. This can be frustrating for users who have specific creative visions for their 3D scene generation, as they cannot directly manipulate the scene's features or details to match their preferences.

In this paper, we present a novel 
system 
that can generate 3D scenes while providing users with fine-grained control over the creation process (see Fig.~\ref{fig:teaser}). 
Despite the existence of 3D generative models~\cite{Zhou_2021_ICCV,chen2019text2shape,cai2020learning}, 
the availability of large-scale 3D datasets required for their training is still limited. Motivated by prior works~\cite{fridman2023scenescape,lin2022magic3d}, we instead rely on 2D diffusion models~\cite{Rombach_2022_CVPR} that have been pre-trained on a large number of 2D images. 
For 3D scene generation, 
we 
use 3D meshes as an intermediary proxy to merge individual 2D images into a unified representation. 

Our system builds upon a generative RGB-D fusion method. 
Specifically, we begin by obtaining an input image from the user or generating one using 2D diffusion. We then utilize a monocular depth estimator~\cite{bhat2023zoedepth} to estimate the underlying geometry of the image and unproject it into 3D space to generate an initial mesh. After transforming the virtual camera to a new viewpoint, we render the mesh and apply 2D diffusion to inpaint holes and outpaint for new content. To ensure seamless integration of the generated content with the existing mesh, we estimate the depth of the image from that viewpoint and perform boundary-aware depth alignment. We then fuse the new mesh with the existing one in 3D space. The above process is repeated iteratively until we obtain a satisfactory complete 3D structure. However, outdoor scenes often pose challenges as 3D meshes cannot handle dramatic depth discontinuities well. To address this issue, we propose to model remote content (e.g., sky) separately with an environment map. This leads to more realistic outdoor scene representation.

To provide users with fine-grained control over the creation process, we develop 
a 3D creator interface 
that enables users to actively participate in the 3D scene creation process.
Our interface offers several advantages. First, users can manipulate the virtual camera to any viewpoint and customize camera trajectories to create personalized 3D scenes. Second, users can adjust the size of the selection box to generate local content, and try different random seeds to generate various results. 
Third, inspired by ControlNet~\cite{zhang2023adding}, we adopt a neural network structure to control diffusion models by adding extra conditions such as user scribbles, semantic segmentation maps, depth, and other information to achieve fine-grained control over the generation process. 
Finally, we introduce a neural rendering interface and incorporate Neural Radiance Fields (NeRFs)~\cite{mildenhall2020nerf,tancik2023nerfstudio} into our 
system, 
allowing users to create a radiance field of their scene online and navigate the entire scene. 
Users can also create camera trajectories to render videos according to their preferences.

In summary, our main contributions are:
\begin{itemize}
    \item 
    We present a new interactive system to generate and render customizable 3D scenes with user control. To this end, we introduce a 3D creator interface and a neural rendering interface.
    \item Our proposed boundary-aware depth alignment allows for the seamless integration of 3D meshes. To better handle outdoor scenes, we propose to model remote content with an environment map rather than 3D meshes.
    \item 
    We achieve interactive 3D scene generation with precise controllability.
\end{itemize}

\section{Related Work}
\paragraph{3D-aware image synthesis.}
Various 3D-GAN based methods~\cite{schwarz2020graf,chan2021pi,chan2022efficient,niemeyer2021giraffe} have been proposed to combine neural scene representations with 2D generative models for 3D-aware image synthesis, enabling direct camera control. While these methods have demonstrated impressive results on the problem of generating single objects such as cars or faces, they are challenging to apply to large and diverse scenes. To extend 3D-aware image synthesis from single objects to completely unconstrained 3D scenes, several recent  works~\cite{Shih3DP20,jampani2021slide,zhou2018stereo,devries2021unconstrained,bautista2022gaudi} have been proposed. For example, GSN~\cite{devries2021unconstrained} proposes to break the radiance field into a grid of local radiance fields and collectively represent a scene by conditioning it on a 2D grid of floorplan latent codes. 
Bautista et al.~\cite{bautista2022gaudi} present GAUDI, where they first optimize a latent representation that disentangles radiance fields and camera poses, and then use the disentangled latent representation to learn a generative model. This allows for both unconditional and conditional generation of 3D scenes. 
However, these methods usually have a significant demand for extensive training and large-scale training data, limiting their generalization to only specific domains. Instead, our objective is to generate diverse 3D scenes.

\paragraph{Perpetual view generation.}
Perpetual view generation~\cite{kaneva2010infinite,liu2021infinite} refers to the process of generating a continuous video sequence that corresponds to an arbitrary camera trajectory, using only a single image of the scene as input. 
Different kinds of methods have been explored in the literature. One line of research~\cite{wiles2020synsin,ren2022look,koh2021pathdreamer,tseng2023consistent} has focused on synthesizing indoor scenes with controllable camera trajectories. Motivated by Liu et al.~\cite{liu2021infinite}, recent works such as InfNat-Zero~\cite{li2022infinitenature} and DiffDreamer~\cite{cai2022diffdreamer} aim at synthesizing flythrough videos of natural landscapes along long camera trajectories. Yet, due to their per-frame generation framework and the lack of underlying scene representations, these methods may suffer from issues such as domain drifting and inconsistent novel views. Recent studies~\cite{chai2023persistent,chen2023scenedreamer} learn a generative model for unconditional synthesis of unbounded 3D nature scenes with a persistent 3D scene representation. Although these methods are capable of producing view-consistent flythrough videos, they necessitate significant training on large-scale datasets and are restricted to a specific domain, e.g., landscapes. On the contrary, our system can generate diverse 3D scenes without the need for large-scale training.

\begin{figure*}[t]
    \centering
    \includegraphics[width=0.9\textwidth]{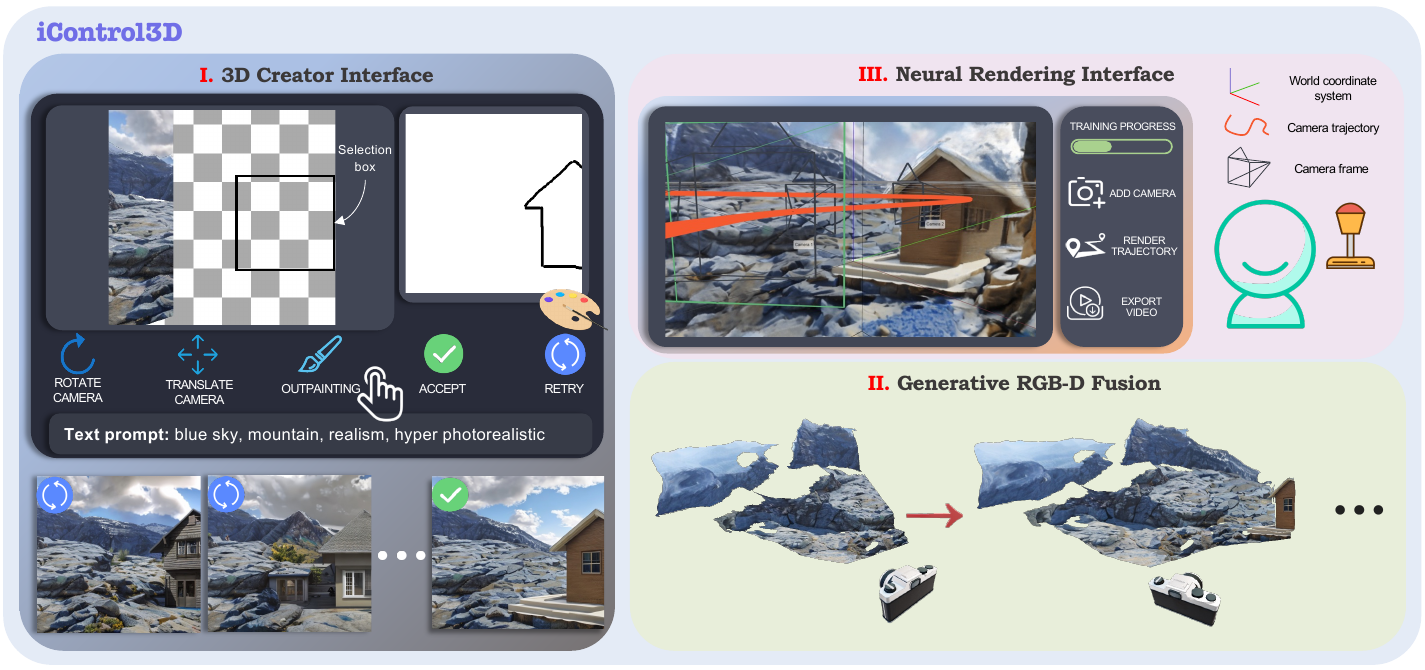}
    \caption{
    \textbf{System overview.} 
    (I) Within our 3D creator interface, users are allowed to manipulate the camera to any viewpoint, adjust the size of the selection box to generate local content, and try different random seeds to create a variety of results. Moreover, users can achieve fine-grained control over the generation process by adding extra conditions such as user scribbles; (II) Once the generated result in (I) is accepted by the users, our generative RGB-D fusion module fuses it with the existing mesh. This alternating process between (I) and (II) continues until a satisfactory 3D structure is obtained; (III) After generating 3D scenes, 
    our neural rendering interface
    then builds  
    a radiance field online and enables users to navigate the entire scene. By recording their virtual journey through the scene, users can also produce high-quality videos that showcase the intricacies and beauty of their designs.}
    \label{fig:overview}
\end{figure*}

\paragraph{3D content generation.}
Diffusion models~\cite{sohl2015deep,ho2020denoising,Rombach_2022_CVPR,saharia2022photorealistic,ramesh2022hierarchical,sun20233d} have demonstrated remarkable success in generating highly realistic images and videos. By iteratively applying a series of steps, these models can transform a simple noise distribution into a complex, high-dimensional data distribution, resulting in images and videos that are virtually indistinguishable from real-world data. As diffusion models continue to advance and gain popularity in the 2D domain, researchers are exploring the possibility of using 2D diffusion priors to generate 3D content. Recent works~\cite{poole2022dreamfusion,Jain_2022_CVPR,lin2022magic3d,lee2022understanding,wang2022score,chan2023generative,wang2023prolificdreamer,shi2023mvdream,liu2023one,liu2023zero} have shown promise in text-guided 3D object generation, but challenges remain in generating large-scale 3D structures and textures for entire scenes. Motivated by previous studies in perpetual view generation~\cite{liu2021infinite,li2022infinitenature,cai2022diffdreamer}, Fridman et al.~\cite{fridman2023scenescape} propose SceneScape, a text-driven approach that synthesizes flying-out trajectories of scenes using 2D diffusion. 
However, SceneScape struggles with generating complete 3D scenes. 
Text2Light~\cite{chen2022text2light} introduces a zero-shot text-driven HDR panorama generation framework for creating 3D scenes but fails to impress users with freely moving cameras. 
Concurrently, 
Text2Room~\cite{hollein2023text2room} uses pre-trained 2D text-to-image diffusion models to create textured 3D meshes of indoor scenes but 
offers limited control over the output. 
To bridge this gap, we present a novel 
system 
that can generate and render customizable 3D scenes with user control.

\section{Method}
\label{sec:method}
\subsection{System Overview}
Our goal is to generate diverse 3D scenes while providing users with fine-grained control over the creation process. 
This entails tackling two challenges, i.e., leveraging 2D diffusion priors for consistent 3D scene generation and 
providing users with controllability over the creation process. 
To achieve our goal, we present iControl3D, an interactive system for 3D scene generation with user control. We schematically illustrate our system in Fig.~\ref{fig:overview}.

Our system mainly consists of a generative RGB-D fusion module, a 3D creator interface, and a neural rendering interface. 
Our system begins by obtaining an input image from users or generating one using 2D diffusion~\cite{Rombach_2022_CVPR}, estimating its geometry via a depth estimator~\cite{bhat2023zoedepth}, and generating an initial 3D mesh. We then render the mesh from different viewpoints, apply inpainting, perform boundary-aware depth alignment, and fuse it with the existing mesh, iteratively refining it until a satisfactory 3D structure is obtained. To handle outdoor scenes with depth discontinuities, we model remote content separately with an environment map, resulting in a more realistic representation.
Unlike previous methods that offer a limited degree of user control, our system presents 
a 3D creator interface 
that enables users to actively participate in the 3D scene creation process. We also incorporate ControlNet~\cite{zhang2023adding}, which can control diffusion models by adding extra conditions, into our interface to provide users with fine-grained control over the synthesized outputs.
After generating 3D scenes, our neural rendering interface then builds a radiance field online and enables users to navigate the entire scene and create camera trajectories to render videos according to their preferences.

\subsection{Generative RGB-D Fusion}
\label{sec:3d-mesh}

\noindent\textbf{Initialization.}
Motivated by previous works~\cite{fridman2023scenescape,lin2022magic3d}, we leverage 2D diffusion models~\cite{Rombach_2022_CVPR} that have been pre-trained on a large number of 2D images. 
Our system starts by obtaining an input image $\bm{I}_{0}$ from users or generating one using 2D diffusion. Formally, let $\mathcal{G}$ be a pre-trained 2D diffusion model. We then can generate the input image $\bm{I}_{0}$ using 2D diffusion model $\mathcal{G}$:
\begin{equation}
    \bm{I}_{0} = \mathcal{G}\left(T, \bm{z}\right),
\end{equation}
where $T$ is a text prompt and $\bm{z}$ represents additional conditions, e.g., user scribbles, semantic segmentation maps, and depth maps. 
It is worth noting that 2D diffusion models can only generate independent 2D images without any 3D structural relationship between them. Hence, relying solely on 2D diffusion models is insufficient to create a unified 3D scene. 
Inspired by prior works~\cite{hollein2023text2room}, we leverage 3D meshes as an intermediary proxy to merge individual 2D images generated by 2D diffusion models into a unified 3D scene representation. 
To this end, we utilize an off-the-shelf monocular depth estimator~\cite{bhat2023zoedepth} to estimate the underlying geometry of the input image. 
After that, we proceed to unproject the input image into an initial 3D mesh $\mathcal{M}_{0} = \left(\mathcal{V}, \mathcal{F}, \mathcal{C}\right)$ using depth values, where $\mathcal{V} = \{ v_{i} \}_{i=1}^{N}$ is the set of $N$ vertices, $\mathcal{F} = \{ f_{i} \}_{i=1}^{F}$ is the set of $F$ faces with each connecting three vertices, and $\mathcal{C} = \{ c_{i} \}_{i=1}^{N}$ are the color vectors attached on vertices. 

\noindent\textbf{\noindent\textbf{Mesh projection and inpainting.}}
We now have the initial 3D mesh $\mathcal{M}_{0}$. 
Our next step is to build up the scene iteratively. To do this, we generate new content from previously unobserved viewpoints. Specifically, we first render the mesh in the target camera pose $\mathbf{P}_{t+1}$:
\begin{equation}
    \hat{\bm{I}}_{t+1}, \hat{\bm{D}}_{t+1}, \hat{\bm{m}}_{t+1} = \Pi\left(\mathcal{M}_{t}, \mathbf{P}_{t+1}\right).
\end{equation}
The mesh renderer $\Pi$~\cite{ravi2020accelerating} produces the rendered image $\hat{\bm{I}}_{t+1}$, the rendered depth $\hat{\bm{D}}_{t+1}$ and the rendered mask $\hat{\bm{m}}_{t+1}$ indicating the visible regions of the mesh in the rendered image, 
where pixels corresponding to visible and invisible parts of the mesh are set to 1 and 0, respectively.
To create new content, the 2D diffusion model $\mathcal{G}$ is employed to inpaint missing pixels via
\begin{equation}
    \bm{I}_{t+1} = \mathcal{G}\left(\hat{\bm{I}}_{t+1}, {\sim}\hat{\bm{m}}_{t+1}, T, \bm{z}\right),
\end{equation}
where ${\sim}\hat{\bm{m}}_{t+1}$ is the inverted mask used to guide the diffusion model by highlighting the areas of the image that should be inpainted. 

\begin{figure}[t]
    \centering
    \includegraphics[width=0.47\textwidth]{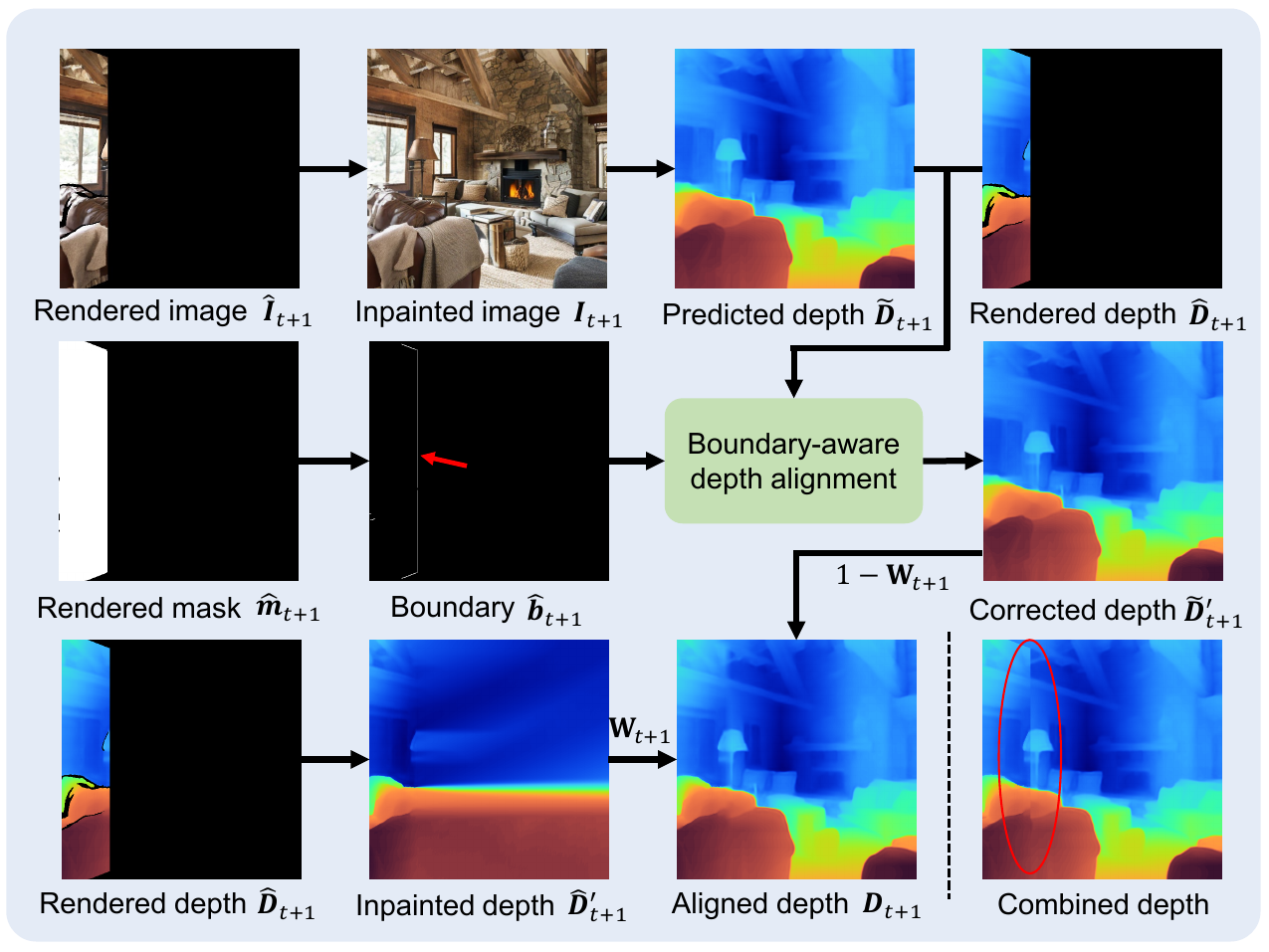}
    \caption{
    \textbf{Boundary-aware depth alignment.} 
    Directly combining the rendered depth $\hat{\bm{D}}_{t+1}$ and the predicted depth $\tilde{\bm{D}}_{t+1}$ leads to abrupt transitions in the combined depth while our boundary-aware depth alignment ensures a more seamless depth fusion. }
    \label{fig:depth-alignment}
\end{figure}

\noindent\textbf{\noindent\textbf{Boundary-aware depth alignment.}}
Likewise, we then employ the depth estimator to predict the underlying geometry of $\bm{I}_{t+1}$, denoted as $\tilde{\bm{D}}_{t+1}$. 
It should be noted that the depth of shared regions between the predicted depth map $\tilde{\bm{D}}_{t+1}$ and the rendered depth map $\hat{\bm{D}}_{t+1}$ may differ. 
To ensure seamless integration of the generated content with the existing mesh, it is intuitive to align the depth such that similar regions in a scene are placed at a similar depth as much as possible. This can help to avoid abrupt transitions at the boundaries between the generated content and the existing mesh. SceneScape~\cite{fridman2023scenescape} utilizes an online test-time training technique to promote the predicted depth map of the current frame to be in line with the geometric structure of the synthesized scene. However, this technique requires a certain amount of time to achieve depth alignment, making it unsuitable for real-time applications. 

To this end, we propose boundary-aware depth alignment. The rationale behind incorporating boundary-aware depth alignment is to minimize the time needed for depth alignment, rendering it suitable for real-time 3D scene generation. This eliminates the necessity of prolonged waiting periods before progressing to the next step in scene creation. As shown in Fig.~\ref{fig:depth-alignment}, we first obtain the boundary $\hat{\bm{b}}_{t+1}$ of the rendered mask $\hat{\bm{m}}_{t+1}$ via:
\begin{equation}
    \hat{\bm{b}}_{t+1} = \left({\sim}\hat{\bm{m}}_{t+1} \oplus \bm{B}\right) \cap \hat{\bm{m}}_{t+1},
\end{equation}
where $\oplus$ and $\cap$ respectively denote the dilation and intersection operation, while $\bm{B}$ represents a structuring element used for the dilation operation. Inspired by Liu et al.~\cite{liu2021infinite}, we then conduct our boundary-aware depth alignment by solving the following 
least squares problem:
\begin{equation}
    \min _{\alpha, \beta}\left\|\hat{\bm{b}}_{t+1} \odot\left(\frac{\alpha}{\tilde{\bm{D}}_{t+1}}+\beta-\frac{1}{\hat{\bm{D}}_{t+1}}\right)\right\|^2,
\end{equation}
where $\odot$ denotes the Hadamard (i.e., element-wise) product. Intuitively, this optimization attempts to scale and shift the predicted disparity using $\alpha$ and $\beta$, so that the aligned disparity of the boundary region matches the rendered disparity. After obtaining the optimal scale and shift parameters, we can use them to compute the corrected depth $\tilde{\bm{D}}_{t+1}^{'}$ as follows:
\begin{equation}
    \tilde{\bm{D}}_{t+1}^{'} = \frac{1}{\alpha / \tilde{\bm{D}}_{t+1} + \beta}.
\end{equation}
To ensure a smoother depth transition, we further propose a depth blending technique. Specifically, we first inpaint the rendered depth $
\hat{\bm{D}}_{t+1}$ with the Navier-Stokes inpainting algorithm~\cite{bertalmio2001navier}, resulting in the inpainted depth $\hat{\bm{D}}_{t+1}^{'}$. Next, we blend the corrected depth $\tilde{\bm{D}}_{t+1}^{'}$ with the inpainted depth $\hat{\bm{D}}_{t+1}^{'}$ to compute the aligned depth $\bm{D}_{t+1}$ as follows:
\begin{equation}
\bm{D}_{t+1} = \mathbf{W}_{t+1} \cdot \hat{\bm{D}}_{t+1}^{'} + (1 - \mathbf{W}_{t+1}) \cdot \tilde{\bm{D}}_{t+1}^{'},
\end{equation}
where the weight map $\mathbf{W}_{t+1}$ is obtained by applying Gaussian blur to $\hat{\bm{m}}_{t+1}$.

\noindent\textbf{\noindent\textbf{Mesh fusion.}}
Given the aligned depth map $\bm{D}_{t+1}$, our next step is to generate a new 3D mesh representation of the scene $\hat{\mathcal{M}}_{t+1}$ from the 2D image $\bm{I}_{t+1}$ and fuse it with the existing mesh $\mathcal{M}_{t}$. 
We first unproject the image pixels into 3D space using the camera intrinsic matrix and target camera pose $\mathbf{P}_{t+1}$. 
Once we have a set of 3D points representing the scene, we follow Höllein et al.~\cite{hollein2023text2room} and 
use a triangulation scheme to construct a mesh representation $\hat{\mathcal{M}}_{t+1}$. This scheme involves connecting each set of four neighboring points in a regular grid pattern to form two triangles. 

To fuse the new 3D mesh $\hat{\mathcal{M}}_{t+1}$ with the existing mesh $\mathcal{M}_{t}$, 
we extend the triangulation scheme at the edges of the inpainting mask ${\sim}\hat{\bm{m}}_{t+1}$ to connect these faces with their neighboring faces from the existing mesh $\mathcal{M}_{t}$. This process results in the fusion of the two meshes, producing the final mesh $\mathcal{M}_{t+1}$. 

\noindent\textbf{\noindent\textbf{Environment map modeling.}}
We can repeat the aforementioned process iteratively until we achieve a satisfactory 3D structure.
However, outdoor scenes may pose a challenge as 3D meshes struggle to handle dramatic depth discontinuities, e.g., between the sky and ground. These discontinuities often lead to flawed structures or large holes in the reconstructed mesh, leading to visible artifacts in the fusion of the new 3D mesh with the existing one. To this end, we propose to model remote content separately with an environment map. Specifically, we assume that remote content has an infinite depth and can be represented as a texture on a sphere surrounding the scene. To embed the remote region into the environment map, for each $\bm{I}_{t+1}$, we use SAM~\cite{kirillov2023segment} to segment the remote region in the image $\bm{I}_{t+1}$ and map each pixel in the segmented region to a point on the surface of a sphere using inverse equirectangular projection. When we change to the next viewpoint, we first obtain the remote content from the environment map, followed by the mesh projection. This allows for accurate rendering of remote regions in subsequent steps, bypassing the issue of depth discontinuities between remote content and foreground.

\subsection{3D Creator Interface}
\label{sec:3d-creator-interface}

Current methods~\cite{hollein2023text2room,fridman2023scenescape,chen2022text2light,chen2023scenedreamer,chai2023persistent,ren2022look} provide limited control over the synthesis process as they only allow for text and predefined camera trajectories as input. This can be frustrating for users who have specific creative visions or requirements for their 3D scene generation, as they cannot directly manipulate the scene's features or details to match their preferences. To overcome this limitation, we introduce a 3D creator interface as a key component of our system. 
The interface provides a user-friendly and intuitive way for users to actively participate in the 3D scene creation process. Our 3D creator interface offers several advantages (see Fig.~\ref{fig:teaser}). One of the most notable features of our interface is the ability for users to adjust the size of the selection box, allowing them to generate local content and try different random seeds to create a variety of results. This feature gives users the ability to select the best output that matches their creative vision. The virtual camera module is another highlight of our interface. It allows users to manipulate the camera to any viewpoint and customize camera trajectories, providing a personalized experience for creating 3D scenes.

\noindent\textbf{\noindent\textbf{Fine-grained control.}}
Our objective is to provide users with not only full control over the 3D scene creation process but also fine-grained control to achieve their desired level of detail and customization.
Inspired by ControlNet~\cite{zhang2023adding}, we adopt a neural network structure to control diffusion models, which allows users to achieve fine-grained control over the generation process by adding extra conditions such as user scribbles, semantic segmentation maps, depth, and other information. This feature enables users to create more complex and detailed 3D scenes.

\subsection{Neural Rendering Interface}
\label{sec:neural-rendering-interface}

Our generative RGB-D fusion module uses 3D meshes as an intermediary proxy to merge individual 2D images into a unified 3D scene representation. 
However, 
we do not employ hole-filling and smoothing techniques as used in previous works~\cite{hollein2023text2room}. This is because we empirically find that iterative mesh reconstruction often leads to unavoidable artifacts in 3D meshes. We instead propose to leverage the 2D diffusion-generated images, which are usually visually pleasing. These images have shared a 3D structural relationship due to our generative RGB-D fusion module. 
We, therefore, introduce a neural rendering interface and integrate Neural Radiance Fields~\cite{mildenhall2020nerf,tancik2023nerfstudio} into our system to further smooth the artifacts shown in 3D meshes.
We train a neural radiance field using the 2D diffusion-generated images and their corresponding poses. 
This enables users to create a radiance field of their scene online and navigate the entire scene during training and after training. Our neural rendering interface offers users an immersive way to explore their 3D creations and produce customized videos.

\setlength{\tabcolsep}{0\linewidth}
\begin{table}[t]
\footnotesize
    \centering
    \caption{
    \textbf{Quantitative comparisons.} 
    We show that our system outperforms all baselines in terms of both the Inception Score (IS)~\cite{salimans2016improved} and CLIP Score (CS)~\cite{radford2021learning}. 
    }
    \begin{tabular}{C{0.5\linewidth}C{0.25\linewidth}C{0.25\linewidth}}
    \toprule
     Method &IS $\uparrow$&CS $\uparrow$\\
    \midrule 
    LOR~\cite{ren2022look} & 1.77   & 20.96\\ 
    SceneDreamer~\cite{chen2023scenedreamer} & 1.35   & 21.35\\ 
    Persistent Nature~\cite{chai2023persistent} & 1.33   & 28.20\\ 
    Text2Room~\cite{hollein2023text2room} & 2.57   & 28.50\\ 
    \rowcolor{LightGray}
    Ours  & \textbf{2.63} & \textbf{29.77}\\ 
    \bottomrule
    \end{tabular}
    \label{tab:quantitative}
\end{table}

\section{Experiments}

\label{sec:exp}
In this section, we present a comprehensive evaluation of our system on a diverse range of indoor and outdoor scenes and compare its performance with state-of-the-art methods both quantitatively and qualitatively. Additionally, we conduct a user study to better evaluate the effectiveness of our system. Finally, we perform an ablation study to justify our design choices. 

\subsection{Baselines}
In our experiments, we primarily compare ours against four representative works including LOR~\cite{ren2022look}, SceneDreamer~\cite{chen2023scenedreamer}, Persistent Nature~\cite{chai2023persistent}, and Text2Room~\cite{hollein2023text2room}.
Specifically, LOR~\cite{ren2022look} is an autoregressive method that can generate long-term 3D indoor scene video from a single image but presents challenges when it comes to generating consistent 3D structures and textures on a scene-scale level. SceneDreamer~\cite{chen2023scenedreamer} and Persistent Nature~\cite{chai2023persistent} learn a generative model for unconditional synthesis of unbounded 3D nature scenes with a persistent 3D scene representation, but necessitate significant training on large-scale datasets and are restricted to a specific domain. Text2Room~\cite{hollein2023text2room} uses pre-trained 2D text-to-image diffusion models to create textured 3D meshes of indoor scenes but lacks fine-grained control over the synthesis process.

\subsection{Results}
\label{sec:results}

\noindent\textbf{\noindent\textbf{Evaluation metrics.}}
To evaluate our system, we utilize Inception Score~\cite{salimans2016improved} and CLIP Score~\cite{radford2021learning} as our evaluation metrics. A higher Inception Score indicates that the generated images have both high quality and diversity, whereas a higher CLIP Score signifies a greater similarity between the generated image and the given text prompt.

\noindent\textbf{\noindent\textbf{Quantitative comparisons.}}
We adopt $21$ scene settings, including $6$ challenging outdoor settings such as ``mountain'' and ``garden'', and $15$ indoor settings such as ``living room'' and ``spaceship'', and randomly generate outdoor scenes twice and indoor scenes once, resulting in $12$ outdoor scenes and $15$ indoor scenes. 
Since our focus is not on achieving complete mesh reconstruction, we instead render $200$ images to compute both Inception Score and CLIP Score for each scene. 
In our evaluation, we closely follow Text2Room. It employs $20$ different trajectories for method evaluation, generating $60$ images from novel viewpoints for each scene to calculate 2D metrics. Likewise, we adopt $21$ scene settings, totaling $27$ scenes for each method, and generate $200$ images for each scene to compute both the IS and CS. Therefore, our evaluation scale aligns with that of Text2Room.
As shown in Table~\ref{tab:quantitative}, our system outperforms existing baselines, 
which indicates that our system produces high-quality and diverse images across different scene settings.

\begin{figure}[t]
    \centering
    \includegraphics[width=0.47\textwidth]{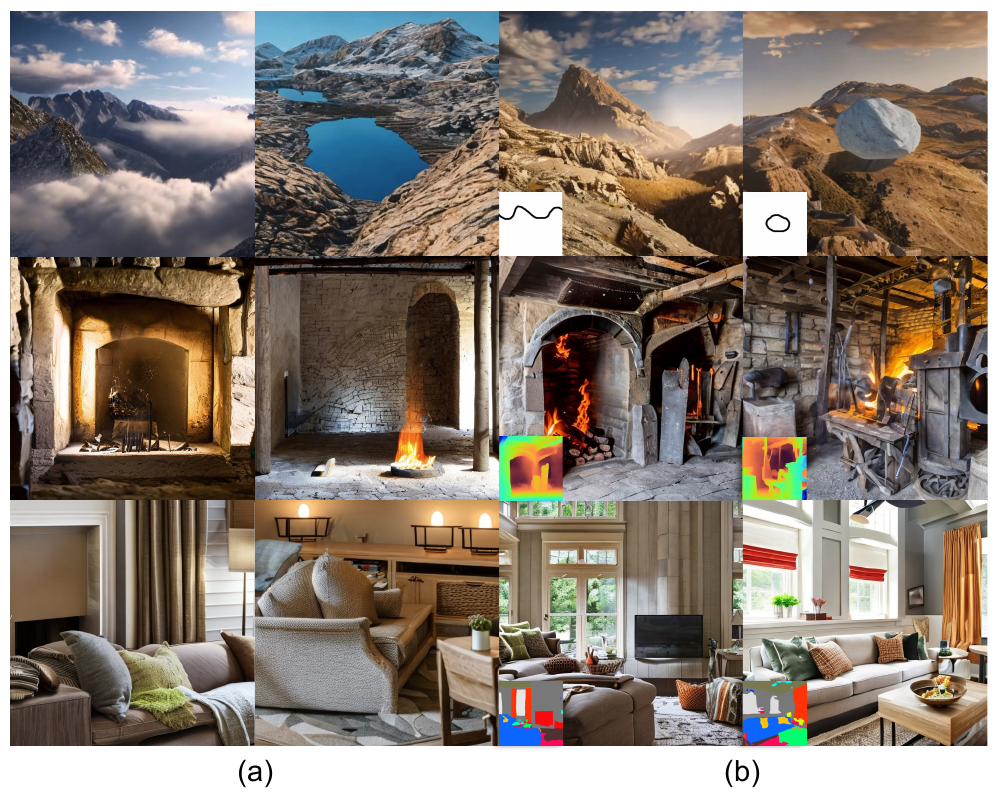}
    \caption{
    \textbf{Fine-grained control.} 
    Compared to (a) current text-driven methods~\cite{hollein2023text2room,fridman2023scenescape}, (b) our system can achieve fine-grained control over the output by adding extra conditions such as scribbles, depth, and semantic segmentation maps. }
    \label{fig:fine-grained-control}
\end{figure}

\begin{figure*}[t]
    \centering
    \includegraphics[width=0.78\textwidth]{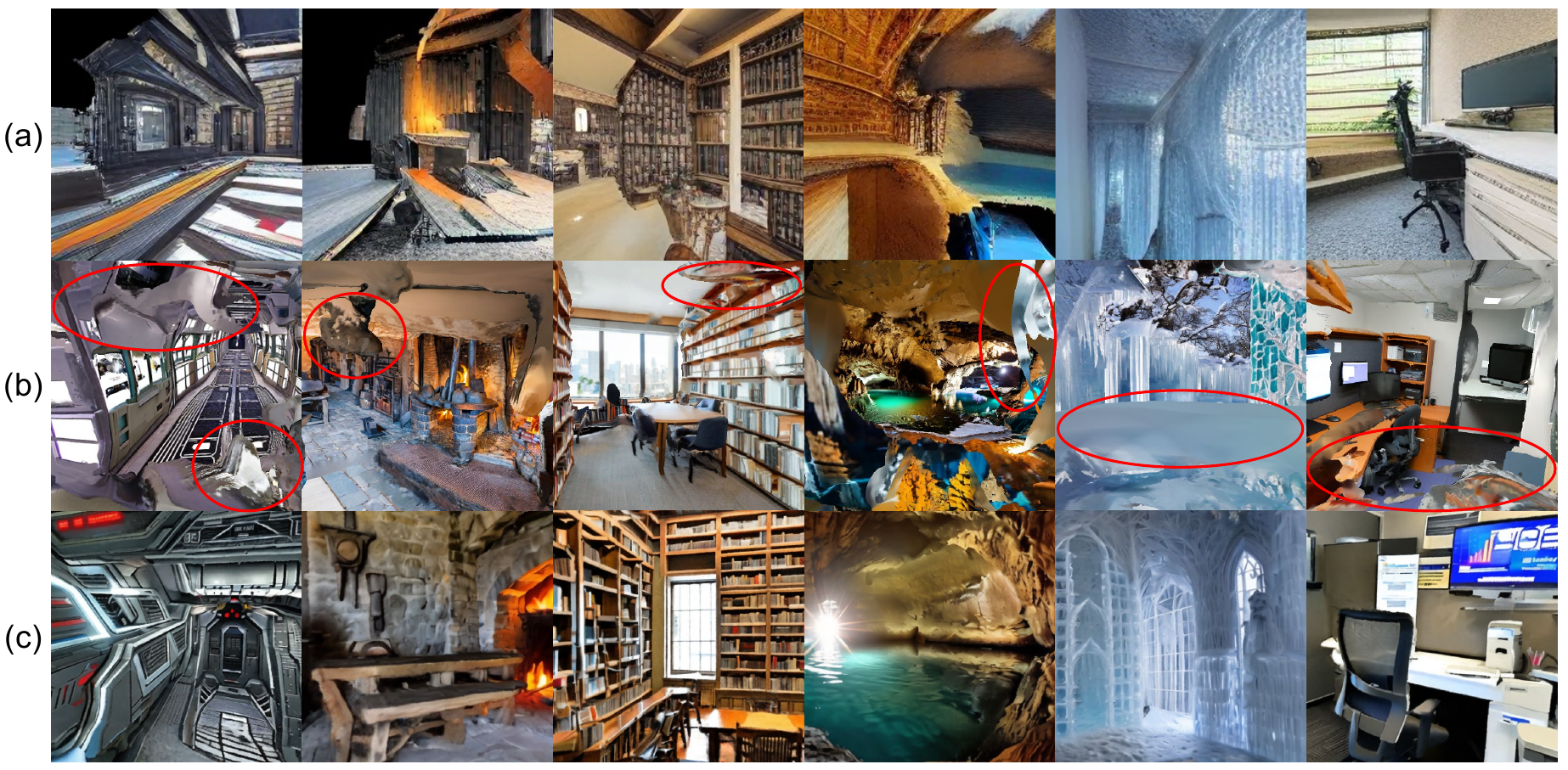}
    \caption{
    \textbf{Qualitative comparison on indoor scenes.} 
    Here we present the qualitative results of indoor scenes, displayed alternately from left to right. 
    The scenes, in sequence, are ``spaceship'', ``forge'', ``library'', ``cave'', ``ice castle'', and ``small office''. 
    (a) LOR~\cite{ren2022look}, (b) Text2Room~\cite{hollein2023text2room}, and (c) ours. 
    As can be seen, (a) LOR~\cite{ren2022look} is prone to domain drifting and a decline in output quality. Although (b) Text2Room~\cite{hollein2023text2room} performs well on indoor scenes, it 
    often produces over-smoothed artifacts in the reconstructions. In contrast, (c) our system presents diverse and photo-realistic results.
    }
    \label{fig:qualitative-indoor}
\end{figure*}

\begin{figure*}[t]
    \centering
    \includegraphics[width=0.785\textwidth]{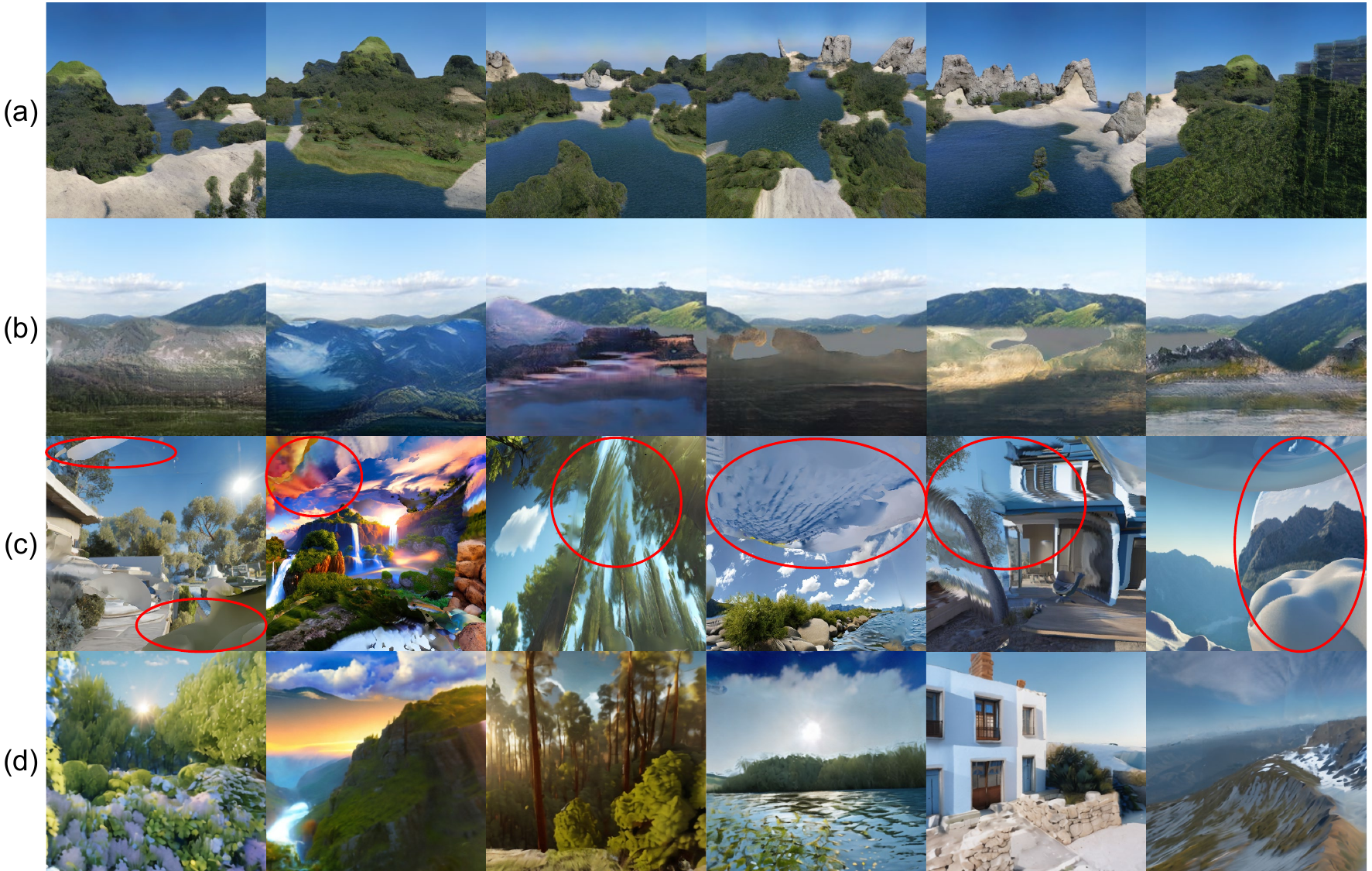}
    \caption{
    \textbf{Qualitative comparison on outdoor scenes.} 
    We present the qualitative results of outdoor scenes, displayed alternately from left to right. 
    The scenes, in sequence, are ``garden'', ``waterfall'', ``forest'', ``river'', ``house'', and ``mountain''. 
    (a) SceneDreamer~\cite{chen2023scenedreamer}, (b) Persistent Nature~\cite{chai2023persistent}, (c) Text2Room~\cite{hollein2023text2room}, and (d) ours. 
    Note that (a) SceneDreamer~\cite{chen2023scenedreamer} and (b) Persistent Nature~\cite{chai2023persistent} require extensive training and are limited to a specific domain, i.e., landscapes. While (c) Text2Room~\cite{hollein2023text2room} can also generate outdoor scenes, it suffers from notable mesh distortions and artifacts. By contrast, (d) our system can generate high-quality and consistent novel views across diverse domains. 
    }
    \label{fig:qualitative-outdoor}
\end{figure*}

\noindent\textbf{\noindent\textbf{Qualitative comparisons.}}
Fig.~\ref{fig:qualitative-indoor} and Fig.~\ref{fig:qualitative-outdoor} present a qualitative comparison between our system and baselines. We showcase randomly extracted novel views of generated scenes. We find that LOR~\cite{ren2022look} exhibits 
the tendency to produce inconsistent novel views and susceptibility to error accumulation. These limitations can lead to domain drifting and a decline in output quality. While SceneDreamer~\cite{chen2023scenedreamer} and Persistent Nature~\cite{chai2023persistent} can synthesize large camera trajectories consistently, they require extensive training and are limited to specific domains such as landscapes. On the other hand, Text2Room~\cite{hollein2023text2room} performs well in indoor scenarios but faces challenges when dealing with outdoor scenes. It also 
often produces over-smoothed regions in the reconstructions. In contrast, 
our system can generate high-quality novel views in both indoor and outdoor scenes. 
In addition, we show in Fig.~\ref{fig:fine-grained-control} that our system can achieve fine-grained control.

\setlength{\tabcolsep}{0\linewidth}
\begin{table}[t]
\footnotesize
    \centering
    \caption{
    \textbf{User study.} 
    We conduct a user study to compare our system against competitive methods. All methods are evaluated on the perceptual quality (PQ) of the imagery and scene diversity (SD). Here we only present pairwise comparison results between ours and baselines.}
    \begin{tabular}{C{0.5\linewidth}C{0.25\linewidth}C{0.25\linewidth}}
    \toprule
    Comparison & PQ $\uparrow$  & SD $\uparrow$\\
    \midrule 
    LOR~\cite{ren2022look} / Ours & 7.6\% / \textbf{92.4\%}   & 1.7\% / \textbf{98.3\%}\\ 
    SceneDreamer~\cite{chen2023scenedreamer} / Ours & 29.5\% / \textbf{70.5\%}  & 11.9\% / \textbf{88.1\%}\\ 
    Persistent Nature~\cite{chai2023persistent} / Ours  & 17.7\% / \textbf{82.3\%}  & 13.8\% / \textbf{86.2\%}\\ 
    Text2Room~\cite{hollein2023text2room} / Ours & 18.6\% / \textbf{81.4\%}   & 25.0\% / \textbf{75.0\%} \\ 
    \bottomrule
    \end{tabular}
    \label{tab:user-study}
\end{table}

\setlength{\tabcolsep}{0\linewidth}
\begin{table}[t]
\footnotesize
    \centering
    \caption{\textbf{Ablation study.} We perform an ablation study on different components of our model to investigate their influence. Each component of our model contributes to the overall performance. }
    \label{tab:ablation}
    \begin{tabular}{C{0.6\linewidth}C{0.2\linewidth}C{0.2\linewidth}}
    \toprule
     Method &IS $\uparrow$&CS $\uparrow$\\
    \midrule 
    w/o boundary-aware depth alignment & 2.19   & 28.37\\ 
    w/o environment map & 2.53   & 29.02\\ 
    \rowcolor{LightGray}
    Full model  & \textbf{2.63} & \textbf{29.77}\\ 
    \bottomrule
    \end{tabular}
\end{table}

\subsection{User Study}
\label{sec:user-study}
To further evaluate the performance of our system, we conduct a user study involving $65$ participants with diverse backgrounds and expertise in the field.
We use different approaches to generate $60$ free-navigating videos of various scenes, respectively. 
To prevent participants from guessing which results are generated by our system during the user study, we randomly present two sets of three videos each time. Both sets consist of three videos generated by randomly selected methods, rather than having one set generated exclusively by our system and the other set by another method. 
Participants are asked to compare two key aspects: the perceptual quality of the imagery and scene diversity. They are invited to choose the method with better perceptual quality and scene diversity, or none if difficult to judge.
We report the results in Table~\ref{tab:user-study}, which points out that our system achieves higher perceptual quality and scene diversity compared to the alternative methods.

\begin{figure}[t]
    \centering
    \includegraphics[width=0.43\textwidth]{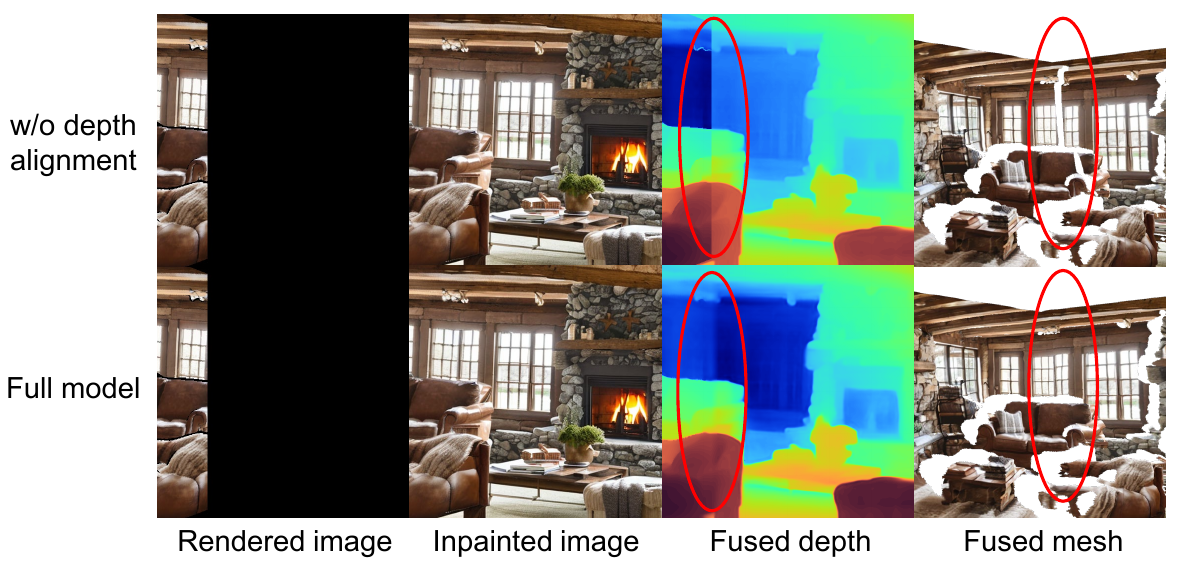}
    \caption{
    \textbf{Effectiveness of boundary-aware depth alignment.} 
    Without boundary-aware depth alignment, the generated mesh may exhibit abrupt transitions 
    at the boundaries between the newly generated content and the existing mesh.
    }
    \label{fig:ablation-depth}
\end{figure}

\subsection{Ablation Study}
\label{sec:ablation-study}
\begin{figure}[t]
    \centering
    \includegraphics[width=0.43\textwidth]{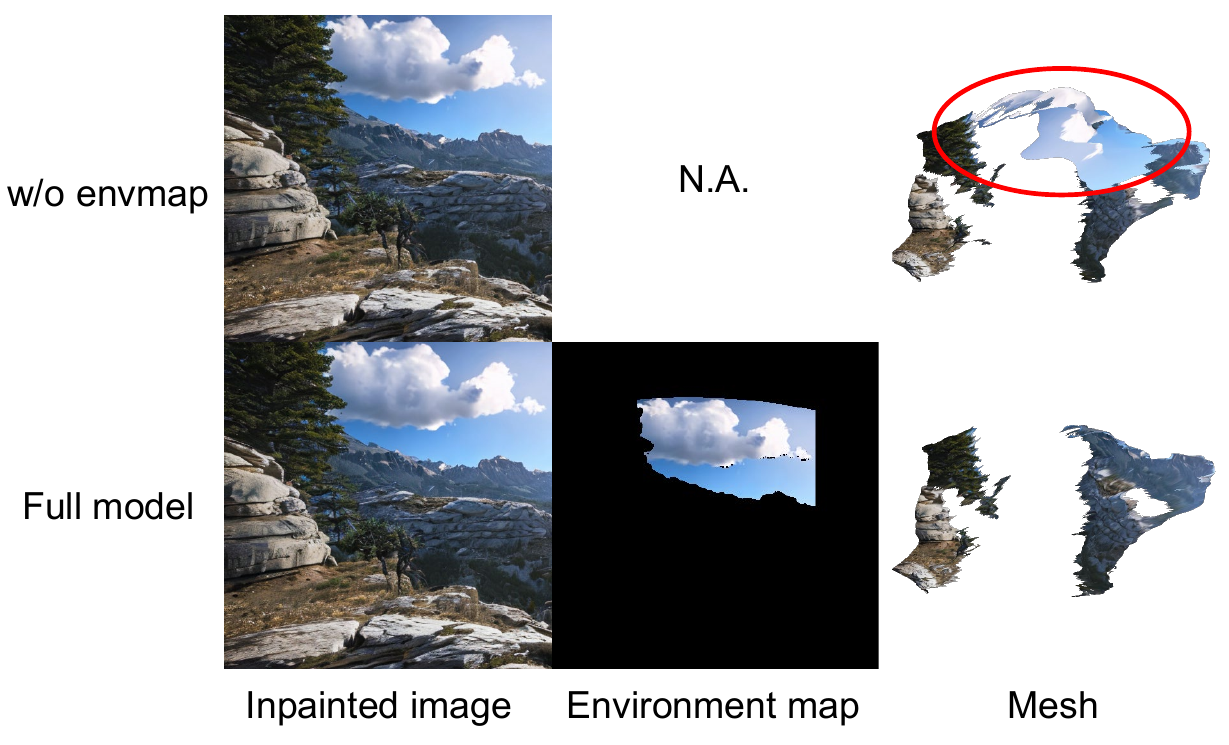}
    \caption{
    \textbf{Effectiveness of environment map modeling.} 
    Without environment map modeling, handling outdoor scenes with 3D meshes becomes challenging due to dramatic depth discontinuities, leading to visible artifacts.
    }
    \label{fig:ablation-envmap}
\end{figure}

To validate the effectiveness of each component of our system, we also conduct an ablation study. We design two variants of our system by removing boundary-aware depth alignment and environment map modeling while keeping the rest of the pipeline intact. 
As shown in Fig.~\ref{fig:ablation-depth} and Fig.~\ref{fig:ablation-envmap}, both boundary-aware depth alignment and environment map modeling contribute to the overall performance. Table~\ref{tab:ablation} also confirms the effect of these components.

\section{Conclusion}

\label{sec:conclusion}

In this paper, we introduce iControl3D, an interactive system for controllable 3D scene generation and rendering. To achieve this, we develop a 3D creator interface to provide users with fine-grained control over the creation process and a neural rendering interface to allow them to navigate the entire scene. We show that our system can generate diverse 3D scenes with user control. We conduct extensive experiments to verify the effectiveness of our system. We hope that our system will inspire and empower users to unleash their creativity and bring their imaginations to life in the world of 3D content creation.

\noindent\textbf{\noindent\textbf{Limitation.}} While our method provides a user-friendly platform for interactive 3D content creation, certain challenges can impact its performance. One such challenge arises when the depth prediction module produces inaccurate geometry based on the input image, or when the segmentation model fails to predict with precision. These issues can compromise the quality of the generated 3D scenes. Moreover, distortions in the 3D meshes can further contribute to inaccuracies and inconsistencies, ultimately affecting the overall realism and quality.


\begin{acks}
This study is supported under the RIE2020 Industry Alignment Fund – Industry Collaboration Projects (IAF-ICP) Funding Initiative, as well as cash and in-kind contribution from the industry partner(s). This work is also supported by the MOE AcRF Tier 2 grant (MOE-T2EP20220-0007).
\end{acks}

\bibliographystyle{ACM-Reference-Format}
\balance
\bibliography{main}


\begin{thebibliography}{58}


\ifx \showCODEN    \undefined \def \showCODEN     #1{\unskip}     \fi
\ifx \showDOI      \undefined \def \showDOI       #1{#1}\fi
\ifx \showISBNx    \undefined \def \showISBNx     #1{\unskip}     \fi
\ifx \showISBNxiii \undefined \def \showISBNxiii  #1{\unskip}     \fi
\ifx \showISSN     \undefined \def \showISSN      #1{\unskip}     \fi
\ifx \showLCCN     \undefined \def \showLCCN      #1{\unskip}     \fi
\ifx \shownote     \undefined \def \shownote      #1{#1}          \fi
\ifx \showarticletitle \undefined \def \showarticletitle #1{#1}   \fi
\ifx \showURL      \undefined \def \showURL       {\relax}        \fi
\providecommand\bibfield[2]{#2}
\providecommand\bibinfo[2]{#2}
\providecommand\natexlab[1]{#1}
\providecommand\showeprint[2][]{arXiv:#2}

\bibitem[Bautista et~al\mbox{.}(2022)]%
        {bautista2022gaudi}
\bibfield{author}{\bibinfo{person}{Miguel~Angel Bautista}, \bibinfo{person}{Pengsheng Guo}, \bibinfo{person}{Samira Abnar}, \bibinfo{person}{Walter Talbott}, \bibinfo{person}{Alexander Toshev}, \bibinfo{person}{Zhuoyuan Chen}, \bibinfo{person}{Laurent Dinh}, \bibinfo{person}{Shuangfei Zhai}, \bibinfo{person}{Hanlin Goh}, \bibinfo{person}{Daniel Ulbricht}, {et~al\mbox{.}}} \bibinfo{year}{2022}\natexlab{}.
\newblock \showarticletitle{Gaudi: A neural architect for immersive 3d scene generation}.
\newblock \bibinfo{journal}{\emph{Advances in Neural Information Processing Systems (NeurIPS)}}  \bibinfo{volume}{35} (\bibinfo{year}{2022}), \bibinfo{pages}{25102--25116}.
\newblock


\bibitem[Bertalmio et~al\mbox{.}(2001)]%
        {bertalmio2001navier}
\bibfield{author}{\bibinfo{person}{Marcelo Bertalmio}, \bibinfo{person}{Andrea~L Bertozzi}, {and} \bibinfo{person}{Guillermo Sapiro}.} \bibinfo{year}{2001}\natexlab{}.
\newblock \showarticletitle{Navier-stokes, fluid dynamics, and image and video inpainting}. In \bibinfo{booktitle}{\emph{Proceedings of the 2001 IEEE Computer Society Conference on Computer Vision and Pattern Recognition. CVPR 2001}}, Vol.~\bibinfo{volume}{1}. IEEE, \bibinfo{pages}{I--I}.
\newblock


\bibitem[Bhat et~al\mbox{.}(2023)]%
        {bhat2023zoedepth}
\bibfield{author}{\bibinfo{person}{Shariq~Farooq Bhat}, \bibinfo{person}{Reiner Birkl}, \bibinfo{person}{Diana Wofk}, \bibinfo{person}{Peter Wonka}, {and} \bibinfo{person}{Matthias M{\"u}ller}.} \bibinfo{year}{2023}\natexlab{}.
\newblock \showarticletitle{Zoedepth: Zero-shot transfer by combining relative and metric depth}.
\newblock \bibinfo{journal}{\emph{arXiv preprint arXiv:2302.12288}} (\bibinfo{year}{2023}).
\newblock


\bibitem[Cai et~al\mbox{.}(2020)]%
        {cai2020learning}
\bibfield{author}{\bibinfo{person}{Ruojin Cai}, \bibinfo{person}{Guandao Yang}, \bibinfo{person}{Hadar Averbuch-Elor}, \bibinfo{person}{Zekun Hao}, \bibinfo{person}{Serge Belongie}, \bibinfo{person}{Noah Snavely}, {and} \bibinfo{person}{Bharath Hariharan}.} \bibinfo{year}{2020}\natexlab{}.
\newblock \showarticletitle{Learning gradient fields for shape generation}. In \bibinfo{booktitle}{\emph{Proceedings of the European Conference on Computer Vision (ECCV)}}. Springer, \bibinfo{pages}{364--381}.
\newblock


\bibitem[Cai et~al\mbox{.}(2022)]%
        {cai2022diffdreamer}
\bibfield{author}{\bibinfo{person}{Shengqu Cai}, \bibinfo{person}{Eric~Ryan Chan}, \bibinfo{person}{Songyou Peng}, \bibinfo{person}{Mohamad Shahbazi}, \bibinfo{person}{Anton Obukhov}, \bibinfo{person}{Luc Van~Gool}, {and} \bibinfo{person}{Gordon Wetzstein}.} \bibinfo{year}{2022}\natexlab{}.
\newblock \showarticletitle{DiffDreamer: Consistent Single-view Perpetual View Generation with Conditional Diffusion Models}.
\newblock \bibinfo{journal}{\emph{arXiv preprint arXiv:2211.12131}} (\bibinfo{year}{2022}).
\newblock


\bibitem[Chai et~al\mbox{.}(2023)]%
        {chai2023persistent}
\bibfield{author}{\bibinfo{person}{Lucy Chai}, \bibinfo{person}{Richard Tucker}, \bibinfo{person}{Zhengqi Li}, \bibinfo{person}{Phillip Isola}, {and} \bibinfo{person}{Noah Snavely}.} \bibinfo{year}{2023}\natexlab{}.
\newblock \showarticletitle{Persistent Nature: A Generative Model of Unbounded 3D Worlds}.
\newblock \bibinfo{journal}{\emph{arXiv preprint arXiv:2303.13515}} (\bibinfo{year}{2023}).
\newblock


\bibitem[Chan et~al\mbox{.}(2022)]%
        {chan2022efficient}
\bibfield{author}{\bibinfo{person}{Eric~R Chan}, \bibinfo{person}{Connor~Z Lin}, \bibinfo{person}{Matthew~A Chan}, \bibinfo{person}{Koki Nagano}, \bibinfo{person}{Boxiao Pan}, \bibinfo{person}{Shalini De~Mello}, \bibinfo{person}{Orazio Gallo}, \bibinfo{person}{Leonidas~J Guibas}, \bibinfo{person}{Jonathan Tremblay}, \bibinfo{person}{Sameh Khamis}, {et~al\mbox{.}}} \bibinfo{year}{2022}\natexlab{}.
\newblock \showarticletitle{Efficient geometry-aware 3D generative adversarial networks}. In \bibinfo{booktitle}{\emph{Proceedings of the IEEE/CVF Conference on Computer Vision and Pattern Recognition (CVPR)}}. \bibinfo{pages}{16123--16133}.
\newblock


\bibitem[Chan et~al\mbox{.}(2021)]%
        {chan2021pi}
\bibfield{author}{\bibinfo{person}{Eric~R Chan}, \bibinfo{person}{Marco Monteiro}, \bibinfo{person}{Petr Kellnhofer}, \bibinfo{person}{Jiajun Wu}, {and} \bibinfo{person}{Gordon Wetzstein}.} \bibinfo{year}{2021}\natexlab{}.
\newblock \showarticletitle{pi-gan: Periodic implicit generative adversarial networks for 3d-aware image synthesis}. In \bibinfo{booktitle}{\emph{Proceedings of the IEEE/CVF Conference on Computer Vision and Pattern Recognition (CVPR)}}. \bibinfo{pages}{5799--5809}.
\newblock


\bibitem[Chan et~al\mbox{.}(2023)]%
        {chan2023generative}
\bibfield{author}{\bibinfo{person}{Eric~R Chan}, \bibinfo{person}{Koki Nagano}, \bibinfo{person}{Matthew~A Chan}, \bibinfo{person}{Alexander~W Bergman}, \bibinfo{person}{Jeong~Joon Park}, \bibinfo{person}{Axel Levy}, \bibinfo{person}{Miika Aittala}, \bibinfo{person}{Shalini De~Mello}, \bibinfo{person}{Tero Karras}, {and} \bibinfo{person}{Gordon Wetzstein}.} \bibinfo{year}{2023}\natexlab{}.
\newblock \showarticletitle{GeNVS: Generative Novel View Synthesis with 3D-Aware Diffusion Models}.
\newblock \bibinfo{journal}{\emph{arXiv preprint arXiv:2304.02602}} (\bibinfo{year}{2023}).
\newblock


\bibitem[Chen et~al\mbox{.}(2019)]%
        {chen2019text2shape}
\bibfield{author}{\bibinfo{person}{Kevin Chen}, \bibinfo{person}{Christopher~B Choy}, \bibinfo{person}{Manolis Savva}, \bibinfo{person}{Angel~X Chang}, \bibinfo{person}{Thomas Funkhouser}, {and} \bibinfo{person}{Silvio Savarese}.} \bibinfo{year}{2019}\natexlab{}.
\newblock \showarticletitle{Text2shape: Generating shapes from natural language by learning joint embeddings}. In \bibinfo{booktitle}{\emph{Proceedings of the Asian Conference on Computer Vision (ACCV)}}. Springer, \bibinfo{pages}{100--116}.
\newblock


\bibitem[Chen et~al\mbox{.}(2022)]%
        {chen2022text2light}
\bibfield{author}{\bibinfo{person}{Zhaoxi Chen}, \bibinfo{person}{Guangcong Wang}, {and} \bibinfo{person}{Ziwei Liu}.} \bibinfo{year}{2022}\natexlab{}.
\newblock \showarticletitle{Text2Light: Zero-Shot Text-Driven HDR Panorama Generation}.
\newblock \bibinfo{journal}{\emph{ACM Transactions on Graphics (TOG)}} \bibinfo{volume}{41}, \bibinfo{number}{6}, Article \bibinfo{articleno}{195} (\bibinfo{year}{2022}), \bibinfo{numpages}{16}~pages.
\newblock


\bibitem[Chen et~al\mbox{.}(2023)]%
        {chen2023scenedreamer}
\bibfield{author}{\bibinfo{person}{Zhaoxi Chen}, \bibinfo{person}{Guangcong Wang}, {and} \bibinfo{person}{Ziwei Liu}.} \bibinfo{year}{2023}\natexlab{}.
\newblock \showarticletitle{Scenedreamer: Unbounded 3d scene generation from 2d image collections}.
\newblock \bibinfo{journal}{\emph{arXiv preprint arXiv:2302.01330}} (\bibinfo{year}{2023}).
\newblock


\bibitem[DeVries et~al\mbox{.}(2021)]%
        {devries2021unconstrained}
\bibfield{author}{\bibinfo{person}{Terrance DeVries}, \bibinfo{person}{Miguel~Angel Bautista}, \bibinfo{person}{Nitish Srivastava}, \bibinfo{person}{Graham~W Taylor}, {and} \bibinfo{person}{Joshua~M Susskind}.} \bibinfo{year}{2021}\natexlab{}.
\newblock \showarticletitle{Unconstrained scene generation with locally conditioned radiance fields}. In \bibinfo{booktitle}{\emph{Proceedings of the IEEE/CVF International Conference on Computer Vision (ICCV)}}. \bibinfo{pages}{14304--14313}.
\newblock


\bibitem[Dhariwal and Nichol(2021)]%
        {dhariwal2021diffusion}
\bibfield{author}{\bibinfo{person}{Prafulla Dhariwal} {and} \bibinfo{person}{Alexander Nichol}.} \bibinfo{year}{2021}\natexlab{}.
\newblock \showarticletitle{Diffusion models beat gans on image synthesis}.
\newblock \bibinfo{journal}{\emph{Advances in Neural Information Processing Systems (NeurIPS)}}  \bibinfo{volume}{34} (\bibinfo{year}{2021}), \bibinfo{pages}{8780--8794}.
\newblock


\bibitem[Fridman et~al\mbox{.}(2023)]%
        {fridman2023scenescape}
\bibfield{author}{\bibinfo{person}{Rafail Fridman}, \bibinfo{person}{Amit Abecasis}, \bibinfo{person}{Yoni Kasten}, {and} \bibinfo{person}{Tali Dekel}.} \bibinfo{year}{2023}\natexlab{}.
\newblock \showarticletitle{Scenescape: Text-driven consistent scene generation}.
\newblock \bibinfo{journal}{\emph{arXiv preprint arXiv:2302.01133}} (\bibinfo{year}{2023}).
\newblock


\bibitem[Ho et~al\mbox{.}(2020)]%
        {ho2020denoising}
\bibfield{author}{\bibinfo{person}{Jonathan Ho}, \bibinfo{person}{Ajay Jain}, {and} \bibinfo{person}{Pieter Abbeel}.} \bibinfo{year}{2020}\natexlab{}.
\newblock \showarticletitle{Denoising diffusion probabilistic models}.
\newblock \bibinfo{journal}{\emph{Advances in Neural Information Processing Systems (NeurIPS)}}  \bibinfo{volume}{33} (\bibinfo{year}{2020}), \bibinfo{pages}{6840--6851}.
\newblock


\bibitem[H{\"o}llein et~al\mbox{.}(2023)]%
        {hollein2023text2room}
\bibfield{author}{\bibinfo{person}{Lukas H{\"o}llein}, \bibinfo{person}{Ang Cao}, \bibinfo{person}{Andrew Owens}, \bibinfo{person}{Justin Johnson}, {and} \bibinfo{person}{Matthias Nie{\ss}ner}.} \bibinfo{year}{2023}\natexlab{}.
\newblock \showarticletitle{Text2Room: Extracting Textured 3D Meshes from 2D Text-to-Image Models}.
\newblock \bibinfo{journal}{\emph{arXiv preprint arXiv:2303.11989}} (\bibinfo{year}{2023}).
\newblock


\bibitem[Jain et~al\mbox{.}(2022)]%
        {Jain_2022_CVPR}
\bibfield{author}{\bibinfo{person}{Ajay Jain}, \bibinfo{person}{Ben Mildenhall}, \bibinfo{person}{Jonathan~T. Barron}, \bibinfo{person}{Pieter Abbeel}, {and} \bibinfo{person}{Ben Poole}.} \bibinfo{year}{2022}\natexlab{}.
\newblock \showarticletitle{Zero-Shot Text-Guided Object Generation With Dream Fields}. In \bibinfo{booktitle}{\emph{Proceedings of the IEEE/CVF Conference on Computer Vision and Pattern Recognition (CVPR)}}. \bibinfo{pages}{867--876}.
\newblock


\bibitem[Jampani et~al\mbox{.}(2021)]%
        {jampani2021slide}
\bibfield{author}{\bibinfo{person}{Varun Jampani}, \bibinfo{person}{Huiwen Chang}, \bibinfo{person}{Kyle Sargent}, \bibinfo{person}{Abhishek Kar}, \bibinfo{person}{Richard Tucker}, \bibinfo{person}{Michael Krainin}, \bibinfo{person}{Dominik Kaeser}, \bibinfo{person}{William~T Freeman}, \bibinfo{person}{David Salesin}, \bibinfo{person}{Brian Curless}, {and} \bibinfo{person}{Ce Liu}.} \bibinfo{year}{2021}\natexlab{}.
\newblock \showarticletitle{{SLIDE}: Single Image 3D Photography with Soft Layering and Depth-aware Inpainting}. In \bibinfo{booktitle}{\emph{Proceedings of the IEEE/CVF International Conference on Computer Vision (ICCV)}}.
\newblock


\bibitem[Kaneva et~al\mbox{.}(2010)]%
        {kaneva2010infinite}
\bibfield{author}{\bibinfo{person}{Biliana Kaneva}, \bibinfo{person}{Josef Sivic}, \bibinfo{person}{Antonio Torralba}, \bibinfo{person}{Shai Avidan}, {and} \bibinfo{person}{William~T Freeman}.} \bibinfo{year}{2010}\natexlab{}.
\newblock \showarticletitle{Infinite images: Creating and exploring a large photorealistic virtual space}.
\newblock \bibinfo{journal}{\emph{Proc. IEEE}} \bibinfo{volume}{98}, \bibinfo{number}{8} (\bibinfo{year}{2010}), \bibinfo{pages}{1391--1407}.
\newblock


\bibitem[Kim et~al\mbox{.}(2023)]%
        {kim2023neuralfield}
\bibfield{author}{\bibinfo{person}{Seung~Wook Kim}, \bibinfo{person}{Bradley Brown}, \bibinfo{person}{Kangxue Yin}, \bibinfo{person}{Karsten Kreis}, \bibinfo{person}{Katja Schwarz}, \bibinfo{person}{Daiqing Li}, \bibinfo{person}{Robin Rombach}, \bibinfo{person}{Antonio Torralba}, {and} \bibinfo{person}{Sanja Fidler}.} \bibinfo{year}{2023}\natexlab{}.
\newblock \showarticletitle{NeuralField-LDM: Scene Generation with Hierarchical Latent Diffusion Models}.
\newblock \bibinfo{journal}{\emph{arXiv preprint arXiv:2304.09787}} (\bibinfo{year}{2023}).
\newblock


\bibitem[Kirillov et~al\mbox{.}(2023)]%
        {kirillov2023segment}
\bibfield{author}{\bibinfo{person}{Alexander Kirillov}, \bibinfo{person}{Eric Mintun}, \bibinfo{person}{Nikhila Ravi}, \bibinfo{person}{Hanzi Mao}, \bibinfo{person}{Chloe Rolland}, \bibinfo{person}{Laura Gustafson}, \bibinfo{person}{Tete Xiao}, \bibinfo{person}{Spencer Whitehead}, \bibinfo{person}{Alexander~C Berg}, \bibinfo{person}{Wan-Yen Lo}, {et~al\mbox{.}}} \bibinfo{year}{2023}\natexlab{}.
\newblock \showarticletitle{Segment anything}.
\newblock \bibinfo{journal}{\emph{arXiv preprint arXiv:2304.02643}} (\bibinfo{year}{2023}).
\newblock


\bibitem[Koh et~al\mbox{.}(2021)]%
        {koh2021pathdreamer}
\bibfield{author}{\bibinfo{person}{Jing~Yu Koh}, \bibinfo{person}{Honglak Lee}, \bibinfo{person}{Yinfei Yang}, \bibinfo{person}{Jason Baldridge}, {and} \bibinfo{person}{Peter Anderson}.} \bibinfo{year}{2021}\natexlab{}.
\newblock \showarticletitle{Pathdreamer: A world model for indoor navigation}. In \bibinfo{booktitle}{\emph{Proceedings of the IEEE/CVF International Conference on Computer Vision (ICCV)}}. \bibinfo{pages}{14738--14748}.
\newblock


\bibitem[Lee and Chang(2022)]%
        {lee2022understanding}
\bibfield{author}{\bibinfo{person}{Han-Hung Lee} {and} \bibinfo{person}{Angel~X Chang}.} \bibinfo{year}{2022}\natexlab{}.
\newblock \showarticletitle{Understanding pure clip guidance for voxel grid nerf models}.
\newblock \bibinfo{journal}{\emph{arXiv preprint arXiv:2209.15172}} (\bibinfo{year}{2022}).
\newblock


\bibitem[Li et~al\mbox{.}(2022)]%
        {li2022infinitenature}
\bibfield{author}{\bibinfo{person}{Zhengqi Li}, \bibinfo{person}{Qianqian Wang}, \bibinfo{person}{Noah Snavely}, {and} \bibinfo{person}{Angjoo Kanazawa}.} \bibinfo{year}{2022}\natexlab{}.
\newblock \showarticletitle{Infinitenature-zero: Learning perpetual view generation of natural scenes from single images}. In \bibinfo{booktitle}{\emph{Proceedings of the European Conference on Computer Vision (ECCV)}}. Springer, \bibinfo{pages}{515--534}.
\newblock


\bibitem[Lin et~al\mbox{.}(2022)]%
        {lin2022magic3d}
\bibfield{author}{\bibinfo{person}{Chen-Hsuan Lin}, \bibinfo{person}{Jun Gao}, \bibinfo{person}{Luming Tang}, \bibinfo{person}{Towaki Takikawa}, \bibinfo{person}{Xiaohui Zeng}, \bibinfo{person}{Xun Huang}, \bibinfo{person}{Karsten Kreis}, \bibinfo{person}{Sanja Fidler}, \bibinfo{person}{Ming-Yu Liu}, {and} \bibinfo{person}{Tsung-Yi Lin}.} \bibinfo{year}{2022}\natexlab{}.
\newblock \showarticletitle{Magic3D: High-Resolution Text-to-3D Content Creation}.
\newblock \bibinfo{journal}{\emph{arXiv preprint arXiv:2211.10440}} (\bibinfo{year}{2022}).
\newblock


\bibitem[Liu et~al\mbox{.}(2021)]%
        {liu2021infinite}
\bibfield{author}{\bibinfo{person}{Andrew Liu}, \bibinfo{person}{Richard Tucker}, \bibinfo{person}{Varun Jampani}, \bibinfo{person}{Ameesh Makadia}, \bibinfo{person}{Noah Snavely}, {and} \bibinfo{person}{Angjoo Kanazawa}.} \bibinfo{year}{2021}\natexlab{}.
\newblock \showarticletitle{Infinite nature: Perpetual view generation of natural scenes from a single image}. In \bibinfo{booktitle}{\emph{Proceedings of the IEEE/CVF International Conference on Computer Vision (ICCV)}}. \bibinfo{pages}{14458--14467}.
\newblock


\bibitem[Liu et~al\mbox{.}(2023b)]%
        {liu2023one}
\bibfield{author}{\bibinfo{person}{Minghua Liu}, \bibinfo{person}{Chao Xu}, \bibinfo{person}{Haian Jin}, \bibinfo{person}{Linghao Chen}, \bibinfo{person}{Zexiang Xu}, \bibinfo{person}{Hao Su}, {et~al\mbox{.}}} \bibinfo{year}{2023}\natexlab{b}.
\newblock \showarticletitle{One-2-3-45: Any single image to 3d mesh in 45 seconds without per-shape optimization}.
\newblock \bibinfo{journal}{\emph{arXiv preprint arXiv:2306.16928}} (\bibinfo{year}{2023}).
\newblock


\bibitem[Liu et~al\mbox{.}(2023a)]%
        {liu2023zero}
\bibfield{author}{\bibinfo{person}{Ruoshi Liu}, \bibinfo{person}{Rundi Wu}, \bibinfo{person}{Basile Van~Hoorick}, \bibinfo{person}{Pavel Tokmakov}, \bibinfo{person}{Sergey Zakharov}, {and} \bibinfo{person}{Carl Vondrick}.} \bibinfo{year}{2023}\natexlab{a}.
\newblock \showarticletitle{Zero-1-to-3: Zero-shot one image to 3d object}. In \bibinfo{booktitle}{\emph{Proceedings of the IEEE/CVF International Conference on Computer Vision}}. \bibinfo{pages}{9298--9309}.
\newblock


\bibitem[Liu et~al\mbox{.}(2023c)]%
        {liu2023grounding}
\bibfield{author}{\bibinfo{person}{Shilong Liu}, \bibinfo{person}{Zhaoyang Zeng}, \bibinfo{person}{Tianhe Ren}, \bibinfo{person}{Feng Li}, \bibinfo{person}{Hao Zhang}, \bibinfo{person}{Jie Yang}, \bibinfo{person}{Chunyuan Li}, \bibinfo{person}{Jianwei Yang}, \bibinfo{person}{Hang Su}, \bibinfo{person}{Jun Zhu}, {et~al\mbox{.}}} \bibinfo{year}{2023}\natexlab{c}.
\newblock \showarticletitle{Grounding dino: Marrying dino with grounded pre-training for open-set object detection}.
\newblock \bibinfo{journal}{\emph{arXiv preprint arXiv:2303.05499}} (\bibinfo{year}{2023}).
\newblock


\bibitem[Mildenhall et~al\mbox{.}(2020)]%
        {mildenhall2020nerf}
\bibfield{author}{\bibinfo{person}{Ben Mildenhall}, \bibinfo{person}{Pratul~P. Srinivasan}, \bibinfo{person}{Matthew Tancik}, \bibinfo{person}{Jonathan~T. Barron}, \bibinfo{person}{Ravi Ramamoorthi}, {and} \bibinfo{person}{Ren Ng}.} \bibinfo{year}{2020}\natexlab{}.
\newblock \showarticletitle{{NeRF}: Representing Scenes as Neural Radiance Fields for View Synthesis}. In \bibinfo{booktitle}{\emph{Proceedings of the European Conference on Computer Vision (ECCV)}}.
\newblock


\bibitem[Niemeyer and Geiger(2021)]%
        {niemeyer2021giraffe}
\bibfield{author}{\bibinfo{person}{Michael Niemeyer} {and} \bibinfo{person}{Andreas Geiger}.} \bibinfo{year}{2021}\natexlab{}.
\newblock \showarticletitle{Giraffe: Representing scenes as compositional generative neural feature fields}. In \bibinfo{booktitle}{\emph{Proceedings of the IEEE/CVF Conference on Computer Vision and Pattern Recognition (CVPR)}}. \bibinfo{pages}{11453--11464}.
\newblock


\bibitem[Poole et~al\mbox{.}(2022)]%
        {poole2022dreamfusion}
\bibfield{author}{\bibinfo{person}{Ben Poole}, \bibinfo{person}{Ajay Jain}, \bibinfo{person}{Jonathan~T Barron}, {and} \bibinfo{person}{Ben Mildenhall}.} \bibinfo{year}{2022}\natexlab{}.
\newblock \showarticletitle{Dreamfusion: Text-to-3d using 2d diffusion}.
\newblock \bibinfo{journal}{\emph{arXiv preprint arXiv:2209.14988}} (\bibinfo{year}{2022}).
\newblock


\bibitem[Radford et~al\mbox{.}(2021)]%
        {radford2021learning}
\bibfield{author}{\bibinfo{person}{Alec Radford}, \bibinfo{person}{Jong~Wook Kim}, \bibinfo{person}{Chris Hallacy}, \bibinfo{person}{Aditya Ramesh}, \bibinfo{person}{Gabriel Goh}, \bibinfo{person}{Sandhini Agarwal}, \bibinfo{person}{Girish Sastry}, \bibinfo{person}{Amanda Askell}, \bibinfo{person}{Pamela Mishkin}, \bibinfo{person}{Jack Clark}, {et~al\mbox{.}}} \bibinfo{year}{2021}\natexlab{}.
\newblock \showarticletitle{Learning transferable visual models from natural language supervision}. In \bibinfo{booktitle}{\emph{International conference on machine learning (ICML)}}. PMLR, \bibinfo{pages}{8748--8763}.
\newblock


\bibitem[Ramesh et~al\mbox{.}(2022)]%
        {ramesh2022hierarchical}
\bibfield{author}{\bibinfo{person}{Aditya Ramesh}, \bibinfo{person}{Prafulla Dhariwal}, \bibinfo{person}{Alex Nichol}, \bibinfo{person}{Casey Chu}, {and} \bibinfo{person}{Mark Chen}.} \bibinfo{year}{2022}\natexlab{}.
\newblock \showarticletitle{Hierarchical text-conditional image generation with clip latents}.
\newblock \bibinfo{journal}{\emph{arXiv preprint arXiv:2204.06125}} (\bibinfo{year}{2022}).
\newblock


\bibitem[Ravi et~al\mbox{.}(2020)]%
        {ravi2020accelerating}
\bibfield{author}{\bibinfo{person}{Nikhila Ravi}, \bibinfo{person}{Jeremy Reizenstein}, \bibinfo{person}{David Novotny}, \bibinfo{person}{Taylor Gordon}, \bibinfo{person}{Wan-Yen Lo}, \bibinfo{person}{Justin Johnson}, {and} \bibinfo{person}{Georgia Gkioxari}.} \bibinfo{year}{2020}\natexlab{}.
\newblock \showarticletitle{Accelerating 3d deep learning with pytorch3d}.
\newblock \bibinfo{journal}{\emph{arXiv preprint arXiv:2007.08501}} (\bibinfo{year}{2020}).
\newblock


\bibitem[Ren and Wang(2022)]%
        {ren2022look}
\bibfield{author}{\bibinfo{person}{Xuanchi Ren} {and} \bibinfo{person}{Xiaolong Wang}.} \bibinfo{year}{2022}\natexlab{}.
\newblock \showarticletitle{Look outside the room: Synthesizing a consistent long-term 3d scene video from a single image}. In \bibinfo{booktitle}{\emph{Proceedings of the IEEE/CVF Conference on Computer Vision and Pattern Recognition (CVPR)}}. \bibinfo{pages}{3563--3573}.
\newblock


\bibitem[Rockwell et~al\mbox{.}(2021)]%
        {rockwell2021pixelsynth}
\bibfield{author}{\bibinfo{person}{Chris Rockwell}, \bibinfo{person}{David~F Fouhey}, {and} \bibinfo{person}{Justin Johnson}.} \bibinfo{year}{2021}\natexlab{}.
\newblock \showarticletitle{Pixelsynth: Generating a 3d-consistent experience from a single image}. In \bibinfo{booktitle}{\emph{Proceedings of the IEEE/CVF International Conference on Computer Vision (ICCV)}}. \bibinfo{pages}{14104--14113}.
\newblock


\bibitem[Rombach et~al\mbox{.}(2022)]%
        {Rombach_2022_CVPR}
\bibfield{author}{\bibinfo{person}{Robin Rombach}, \bibinfo{person}{Andreas Blattmann}, \bibinfo{person}{Dominik Lorenz}, \bibinfo{person}{Patrick Esser}, {and} \bibinfo{person}{Bj\"orn Ommer}.} \bibinfo{year}{2022}\natexlab{}.
\newblock \showarticletitle{High-Resolution Image Synthesis With Latent Diffusion Models}. In \bibinfo{booktitle}{\emph{Proceedings of the IEEE/CVF Conference on Computer Vision and Pattern Recognition (CVPR)}}. \bibinfo{pages}{10684--10695}.
\newblock


\bibitem[Saharia et~al\mbox{.}(2022)]%
        {saharia2022photorealistic}
\bibfield{author}{\bibinfo{person}{Chitwan Saharia}, \bibinfo{person}{William Chan}, \bibinfo{person}{Saurabh Saxena}, \bibinfo{person}{Lala Li}, \bibinfo{person}{Jay Whang}, \bibinfo{person}{Emily~L Denton}, \bibinfo{person}{Kamyar Ghasemipour}, \bibinfo{person}{Raphael Gontijo~Lopes}, \bibinfo{person}{Burcu Karagol~Ayan}, \bibinfo{person}{Tim Salimans}, {et~al\mbox{.}}} \bibinfo{year}{2022}\natexlab{}.
\newblock \showarticletitle{Photorealistic text-to-image diffusion models with deep language understanding}.
\newblock \bibinfo{journal}{\emph{Advances in Neural Information Processing Systems (NeurIPS)}}  \bibinfo{volume}{35} (\bibinfo{year}{2022}), \bibinfo{pages}{36479--36494}.
\newblock


\bibitem[Salimans et~al\mbox{.}(2016)]%
        {salimans2016improved}
\bibfield{author}{\bibinfo{person}{Tim Salimans}, \bibinfo{person}{Ian Goodfellow}, \bibinfo{person}{Wojciech Zaremba}, \bibinfo{person}{Vicki Cheung}, \bibinfo{person}{Alec Radford}, {and} \bibinfo{person}{Xi Chen}.} \bibinfo{year}{2016}\natexlab{}.
\newblock \showarticletitle{Improved techniques for training gans}.
\newblock \bibinfo{journal}{\emph{Advances in Neural Information Processing Systems (NeurIPS)}}  \bibinfo{volume}{29} (\bibinfo{year}{2016}).
\newblock


\bibitem[Schwarz et~al\mbox{.}(2020)]%
        {schwarz2020graf}
\bibfield{author}{\bibinfo{person}{Katja Schwarz}, \bibinfo{person}{Yiyi Liao}, \bibinfo{person}{Michael Niemeyer}, {and} \bibinfo{person}{Andreas Geiger}.} \bibinfo{year}{2020}\natexlab{}.
\newblock \showarticletitle{Graf: Generative radiance fields for 3d-aware image synthesis}.
\newblock \bibinfo{journal}{\emph{Advances in Neural Information Processing Systems (NeurIPS)}}  \bibinfo{volume}{33} (\bibinfo{year}{2020}), \bibinfo{pages}{20154--20166}.
\newblock


\bibitem[Shen et~al\mbox{.}(2022)]%
        {shen2022sgam}
\bibfield{author}{\bibinfo{person}{Yuan Shen}, \bibinfo{person}{Wei-Chiu Ma}, {and} \bibinfo{person}{Shenlong Wang}.} \bibinfo{year}{2022}\natexlab{}.
\newblock \showarticletitle{{SGAM}: Building a Virtual 3D World through Simultaneous Generation and Mapping}. In \bibinfo{booktitle}{\emph{Advances in Neural Information Processing Systems (NeurIPS)}}.
\newblock


\bibitem[Shi et~al\mbox{.}(2023)]%
        {shi2023mvdream}
\bibfield{author}{\bibinfo{person}{Yichun Shi}, \bibinfo{person}{Peng Wang}, \bibinfo{person}{Jianglong Ye}, \bibinfo{person}{Mai Long}, \bibinfo{person}{Kejie Li}, {and} \bibinfo{person}{Xiao Yang}.} \bibinfo{year}{2023}\natexlab{}.
\newblock \showarticletitle{Mvdream: Multi-view diffusion for 3d generation}.
\newblock \bibinfo{journal}{\emph{arXiv preprint arXiv:2308.16512}} (\bibinfo{year}{2023}).
\newblock


\bibitem[Shih et~al\mbox{.}(2020)]%
        {Shih3DP20}
\bibfield{author}{\bibinfo{person}{Meng-Li Shih}, \bibinfo{person}{Shih-Yang Su}, \bibinfo{person}{Johannes Kopf}, {and} \bibinfo{person}{Jia-Bin Huang}.} \bibinfo{year}{2020}\natexlab{}.
\newblock \showarticletitle{3D Photography using Context-aware Layered Depth Inpainting}. In \bibinfo{booktitle}{\emph{Proceedings of the IEEE/CVF Conference on Computer Vision and Pattern Recognition (CVPR)}}.
\newblock


\bibitem[Sohl-Dickstein et~al\mbox{.}(2015)]%
        {sohl2015deep}
\bibfield{author}{\bibinfo{person}{Jascha Sohl-Dickstein}, \bibinfo{person}{Eric Weiss}, \bibinfo{person}{Niru Maheswaranathan}, {and} \bibinfo{person}{Surya Ganguli}.} \bibinfo{year}{2015}\natexlab{}.
\newblock \showarticletitle{Deep unsupervised learning using nonequilibrium thermodynamics}. In \bibinfo{booktitle}{\emph{International Conference on Machine Learning (ICML)}}. PMLR, \bibinfo{pages}{2256--2265}.
\newblock


\bibitem[Sun et~al\mbox{.}(2023)]%
        {sun20233d}
\bibfield{author}{\bibinfo{person}{Chunyi Sun}, \bibinfo{person}{Junlin Han}, \bibinfo{person}{Weijian Deng}, \bibinfo{person}{Xinlong Wang}, \bibinfo{person}{Zishan Qin}, {and} \bibinfo{person}{Stephen Gould}.} \bibinfo{year}{2023}\natexlab{}.
\newblock \showarticletitle{3D-GPT: Procedural 3D Modeling with Large Language Models}.
\newblock \bibinfo{journal}{\emph{arXiv preprint arXiv:2310.12945}} (\bibinfo{year}{2023}).
\newblock


\bibitem[Tancik et~al\mbox{.}(2023)]%
        {tancik2023nerfstudio}
\bibfield{author}{\bibinfo{person}{Matthew Tancik}, \bibinfo{person}{Ethan Weber}, \bibinfo{person}{Evonne Ng}, \bibinfo{person}{Ruilong Li}, \bibinfo{person}{Brent Yi}, \bibinfo{person}{Justin Kerr}, \bibinfo{person}{Terrance Wang}, \bibinfo{person}{Alexander Kristoffersen}, \bibinfo{person}{Jake Austin}, \bibinfo{person}{Kamyar Salahi}, {et~al\mbox{.}}} \bibinfo{year}{2023}\natexlab{}.
\newblock \showarticletitle{Nerfstudio: A modular framework for neural radiance field development}.
\newblock \bibinfo{journal}{\emph{arXiv preprint arXiv:2302.04264}} (\bibinfo{year}{2023}).
\newblock


\bibitem[Tseng et~al\mbox{.}(2023)]%
        {tseng2023consistent}
\bibfield{author}{\bibinfo{person}{Hung-Yu Tseng}, \bibinfo{person}{Qinbo Li}, \bibinfo{person}{Changil Kim}, \bibinfo{person}{Suhib Alsisan}, \bibinfo{person}{Jia-Bin Huang}, {and} \bibinfo{person}{Johannes Kopf}.} \bibinfo{year}{2023}\natexlab{}.
\newblock \showarticletitle{Consistent View Synthesis with Pose-Guided Diffusion Models}.
\newblock \bibinfo{journal}{\emph{arXiv preprint arXiv:2303.17598}} (\bibinfo{year}{2023}).
\newblock


\bibitem[Wang et~al\mbox{.}(2022)]%
        {wang2022score}
\bibfield{author}{\bibinfo{person}{Haochen Wang}, \bibinfo{person}{Xiaodan Du}, \bibinfo{person}{Jiahao Li}, \bibinfo{person}{Raymond~A Yeh}, {and} \bibinfo{person}{Greg Shakhnarovich}.} \bibinfo{year}{2022}\natexlab{}.
\newblock \showarticletitle{Score Jacobian Chaining: Lifting Pretrained 2D Diffusion Models for 3D Generation}.
\newblock \bibinfo{journal}{\emph{arXiv preprint arXiv:2212.00774}} (\bibinfo{year}{2022}).
\newblock


\bibitem[Wang et~al\mbox{.}(2023)]%
        {wang2023prolificdreamer}
\bibfield{author}{\bibinfo{person}{Zhengyi Wang}, \bibinfo{person}{Cheng Lu}, \bibinfo{person}{Yikai Wang}, \bibinfo{person}{Fan Bao}, \bibinfo{person}{Chongxuan Li}, \bibinfo{person}{Hang Su}, {and} \bibinfo{person}{Jun Zhu}.} \bibinfo{year}{2023}\natexlab{}.
\newblock \showarticletitle{ProlificDreamer: High-Fidelity and Diverse Text-to-3D Generation with Variational Score Distillation}.
\newblock \bibinfo{journal}{\emph{arXiv preprint arXiv:2305.16213}} (\bibinfo{year}{2023}).
\newblock


\bibitem[Wiles et~al\mbox{.}(2020)]%
        {wiles2020synsin}
\bibfield{author}{\bibinfo{person}{Olivia Wiles}, \bibinfo{person}{Georgia Gkioxari}, \bibinfo{person}{Richard Szeliski}, {and} \bibinfo{person}{Justin Johnson}.} \bibinfo{year}{2020}\natexlab{}.
\newblock \showarticletitle{Synsin: End-to-end view synthesis from a single image}. In \bibinfo{booktitle}{\emph{Proceedings of the IEEE/CVF Conference on Computer Vision and Pattern Recognition (CVPR)}}. \bibinfo{pages}{7467--7477}.
\newblock


\bibitem[Yu et~al\mbox{.}(2015)]%
        {yu2015clutterpalette}
\bibfield{author}{\bibinfo{person}{Lap-Fai Yu}, \bibinfo{person}{Sai-Kit Yeung}, {and} \bibinfo{person}{Demetri Terzopoulos}.} \bibinfo{year}{2015}\natexlab{}.
\newblock \showarticletitle{The Clutterpalette: An Interactive Tool for Detailing Indoor Scenes}.
\newblock \bibinfo{journal}{\emph{IEEE Transactions on Visualization and Computer Graphics}} \bibinfo{volume}{22}, \bibinfo{number}{2} (\bibinfo{year}{2015}), \bibinfo{pages}{1138--1148}.
\newblock


\bibitem[Zhang and Agrawala(2023)]%
        {zhang2023adding}
\bibfield{author}{\bibinfo{person}{Lvmin Zhang} {and} \bibinfo{person}{Maneesh Agrawala}.} \bibinfo{year}{2023}\natexlab{}.
\newblock \showarticletitle{Adding conditional control to text-to-image diffusion models}.
\newblock \bibinfo{journal}{\emph{arXiv preprint arXiv:2302.05543}} (\bibinfo{year}{2023}).
\newblock


\bibitem[Zhang et~al\mbox{.}(2021)]%
        {zhang2021mageadd}
\bibfield{author}{\bibinfo{person}{Shao-Kui Zhang}, \bibinfo{person}{Yi-Xiao Li}, \bibinfo{person}{Yu He}, \bibinfo{person}{Yong-Liang Yang}, {and} \bibinfo{person}{Song-Hai Zhang}.} \bibinfo{year}{2021}\natexlab{}.
\newblock \showarticletitle{MageAdd: Real-Time Interaction Simulation for Scene Synthesis}. In \bibinfo{booktitle}{\emph{Proceedings of the 29th ACM International Conference on Multimedia (ACM MM)}}. \bibinfo{pages}{965--973}.
\newblock


\bibitem[Zhang et~al\mbox{.}(2023)]%
        {zhang2023scenedirector}
\bibfield{author}{\bibinfo{person}{Shao-Kui Zhang}, \bibinfo{person}{Hou Tam}, \bibinfo{person}{Yike Li}, \bibinfo{person}{Ke-Xin Ren}, \bibinfo{person}{Hongbo Fu}, {and} \bibinfo{person}{Song-Hai Zhang}.} \bibinfo{year}{2023}\natexlab{}.
\newblock \showarticletitle{SceneDirector: Interactive Scene Synthesis by Simultaneously Editing Multiple Objects in Real-Time}.
\newblock \bibinfo{journal}{\emph{IEEE Transactions on Visualization and Computer Graphics}} \bibinfo{volume}{30}, \bibinfo{number}{8} (\bibinfo{year}{2023}), \bibinfo{pages}{4558--4569}.
\newblock


\bibitem[Zhou et~al\mbox{.}(2021)]%
        {Zhou_2021_ICCV}
\bibfield{author}{\bibinfo{person}{Linqi Zhou}, \bibinfo{person}{Yilun Du}, {and} \bibinfo{person}{Jiajun Wu}.} \bibinfo{year}{2021}\natexlab{}.
\newblock \showarticletitle{3D Shape Generation and Completion Through Point-Voxel Diffusion}. In \bibinfo{booktitle}{\emph{Proceedings of the IEEE/CVF International Conference on Computer Vision (ICCV)}}. \bibinfo{pages}{5826--5835}.
\newblock


\bibitem[Zhou et~al\mbox{.}(2018)]%
        {zhou2018stereo}
\bibfield{author}{\bibinfo{person}{Tinghui Zhou}, \bibinfo{person}{Richard Tucker}, \bibinfo{person}{John Flynn}, \bibinfo{person}{Graham Fyffe}, {and} \bibinfo{person}{Noah Snavely}.} \bibinfo{year}{2018}\natexlab{}.
\newblock \showarticletitle{Stereo magnification: learning view synthesis using multiplane images}.
\newblock \bibinfo{journal}{\emph{ACM Transactions on Graphics (TOG)}} \bibinfo{volume}{37}, \bibinfo{number}{4} (\bibinfo{year}{2018}), \bibinfo{pages}{1--12}.
\newblock


\end{thebibliography}

\clearpage
\appendix

\section{Contribution Revisited}
Existing 3D scene generation methods often lack adequate user control, leading to the generation of scenes that may not align with users' preferences. Our primary objective is to overcome this limitation and provide users with enhanced controllability in the creation of 3D scenes. To achieve this goal, we propose an interactive system that seamlessly combines sequential scene generation with user controllability. Besides the interactive UI, one of the challenges is that the depth estimation network may produce depth maps that exhibit inconsistencies in scale between two frames. To solve this, we introduce a novel technique called boundary-aware depth alignment. This approach ensures a smooth integration of 3D meshes. Moreover, to better handle outdoor scenes, we integrate an environment map into our system, further enhancing the overall scene generation process.

Our proposed system differs from previous works in four ways: i) Boundary-aware depth alignment (faster, comparable performance);
ii) Outdoor scene generation (not fully addressed in previous works); iii) Adding finer-grained control to the scene generation process (not fully addressed in previous works); iv) 
Consolidating the final results in neural radiance fields to ameliorate artifacts common to meshes.

\section{Implementation Details}
Our system leverages Stable Diffusion~\cite{Rombach_2022_CVPR}, which has been pre-trained on a large number of 2D images, to generate a diverse range of images. To enable controllable 3D scene generation, we integrate ControlNet~\cite{zhang2023adding} with the pre-trained large diffusion models to support additional input conditions and enhance control over the generated scenes.
For estimating depth maps, we utilize an off-the-shelf monocular depth estimator~\cite{bhat2023zoedepth} to estimate the underlying geometry of the input image. This allows us to predict dense depth maps for in-the-wild photos. The implementation of mesh reconstruction, projection, and fusion is carried out using PyTorch3D~\cite{ravi2020accelerating}.
To segment remote content such as the sky, we first employ Grounding DINO~\cite{liu2023grounding} to detect remote regions based on text inputs. Then, we apply SAM~\cite{kirillov2023segment} for precise segmentation.
Our 3D creator interface is based on an open-source project\footnote{https://github.com/lkwq007/stablediffusion-infinity} and implemented using PyScript and Gradio, providing a user-friendly interface for creating 3D scenes. The neural rendering interface is built on top of Nerfstudio~\cite{tancik2023nerfstudio}, which enables users to navigate the entire scene freely and render customizable videos according to their preferences. We conduct all experiments on a single NVIDIA GeForce RTX 3090 GPU. Our code will be publicly available upon acceptance for academic purposes.

\setlength{\tabcolsep}{0\linewidth}
\begin{table*}[t]
\footnotesize
    \centering
    \caption{
    \textbf{Comparison of ours and relevant works.}
    \textit{Indoor scene}: Designed for handling indoor scenes. 
    \textit{Outdoor scene}: Designed for handling outdoor scenes. 
    \textit{No large-scale training}: Not requiring large-scale training.
    \textit{Radiance field}: If radiance fields are used.
    \textit{Interactive generation}: If interactive generation is supported using an interface.
    \textit{Local generation}: If local generation is supported.
    \textit{Conditional synthesis}: If the synthesis can be conditioned on additional input.
    \textit{Text control}: If the generation can be controlled by text prompts.
    \textit{Fine-grained control}: If having precise control over the generation process, e.g., scribbles.
    }
    \begin{tabular}{C{0.18\linewidth}C{0.08\linewidth}C{0.08\linewidth}C{0.13\linewidth}C{0.09\linewidth}C{0.09\linewidth}C{0.09\linewidth}C{0.09\linewidth}C{0.07\linewidth}C{0.1\linewidth}}
    \toprule
     \multirow{2}{*}{Method} &Indoor&Outdoor& No large-scale&Radiance&Interactive&Local&Conditional&Text&Fine-grained\\
     &scene&scene&training&field&generation&generation&synthesis&control&control\\
    \midrule 
    PixelSynth~\cite{rockwell2021pixelsynth} & \cmark   & \xmark & \xmark & \xmark & \xmark & \xmark & \cmark & \xmark & \xmark\\ 
    InfNat-Zero~\cite{li2022infinitenature} & \xmark   & \cmark  & \xmark & \xmark & \xmark & \xmark& \cmark & \xmark & \xmark\\ 
    LOR~\cite{ren2022look} & \cmark   & \xmark  & \xmark & \xmark & \xmark & \xmark & \cmark & \xmark & \xmark\\ 
    SceneScape~\cite{fridman2023scenescape} & \cmark   & \xmark  & \cmark & \xmark & \xmark & \xmark & \cmark & \cmark & \xmark\\ 
    GSN~\cite{devries2021unconstrained}  & \cmark  & \xmark  & \xmark & \cmark & \xmark & \xmark& \cmark & \xmark & \xmark\\ 
    SGAM~\cite{shen2022sgam} & \cmark   & \cmark  & \xmark & \xmark & \xmark& \xmark & \cmark & \xmark & \xmark\\ 
    SceneDreamer~\cite{chen2023scenedreamer} & \xmark   & \cmark  & \xmark & \cmark & \xmark & \xmark & \xmark & \xmark & \xmark\\ 
    Persistent Nature~\cite{chai2023persistent} & \xmark   & \cmark  & \xmark & \cmark & \xmark & \xmark & \xmark & \xmark & \xmark\\ 
    Text2Room~\cite{hollein2023text2room} & \cmark   & \xmark  & \cmark & \xmark & \xmark & \xmark & \cmark & \cmark & \xmark\\ 
    NF-LDM~\cite{kim2023neuralfield} & \cmark   & \cmark  & \xmark & \cmark & \xmark & \xmark & \cmark & \xmark & \xmark\\ 
    \rowcolor{LightGray}
    Ours  & \cmark & \cmark  & \cmark & \cmark & \cmark & \cmark & \cmark & \cmark & \cmark\\ 
    \bottomrule
    \end{tabular}
    \label{tab:contributions}
\end{table*}

\section{Baselines}
Table~\ref{tab:contributions} presents a comparison of our method with other relevant works. In our experiments, we primarily compare ours against four representative works including LOR~\cite{ren2022look}, SceneDreamer~\cite{chen2023scenedreamer}, Persistent Nature~\cite{chai2023persistent}, and Text2Room~\cite{hollein2023text2room}. Specifically, LOR~\cite{ren2022look} is an autoregressive method that can generate long-term 3D indoor scene video from a single image but presents challenges when it comes to generating consistent 3D structures and textures on a scene-scale level. SceneDreamer~\cite{chen2023scenedreamer} and Persistent Nature~\cite{chai2023persistent} learn a generative model for unconditional synthesis of unbounded 3D nature scenes with a persistent 3D scene representation, but necessitate significant training on large-scale datasets and are restricted to a specific domain. Text2Room~\cite{hollein2023text2room} uses pre-trained 2D text-to-image diffusion models to create textured 3D meshes of indoor scenes but lacks fine-grained control over the synthesis process. We implement these works using the official codes released on GitHub. 

We emphasize that our method distinguishes Text2Room in four significant ways. The most notable difference from Text2Room is the introduction of an interactive system designed to facilitate the comprehensive creation of a 3D scene by the user. The key strength of this system lies in its capacity to empower users with finer control over generated content. It allows for the integration of text with other modalities such as scribbles and semantic segmentation maps, offering users the capability to select specific parts of the scene for focus. Secondly, while Text2Room employs scale-and-shift depth alignment, our method goes a step further by incorporating a depth blending technique around the boundary. This enhancement ensures a smoother depth transition in the generated scenes. Thirdly, Text2Room is limited to handling indoor scenes, whereas our method extends its capabilities to generate outdoor scenes through the incorporation of environment maps. This broadens the scope of scene generation possibilities beyond indoor environments. Furthermore, we integrate Neural Radiance Fields into our system to further smooth the artifacts shown in 3D meshes.

\section{Differences from Interactive Scene Synthesis Frameworks}
Broadly speaking, those interactive scene synthesis frameworks~\cite{yu2015clutterpalette,zhang2023scenedirector,zhang2021mageadd} are tailored to assist modelers in manipulating groups of objects—usually CAD models—by enabling insertion, removal, translation, and rotation to enhance scene complexity. However, these methods primarily focus on indoor scenes and operating known objects. In contrast, our system exhibits versatility by generating diverse results and is not confined to indoor environments alone.

\section{Experiment Details}
In the paper, we provide a comprehensive comparison with LOR~\cite{ren2022look}, SceneDreamer~\cite{chen2023scenedreamer}, Persistent Nature~\cite{chai2023persistent}, and Text2Room~\cite{hollein2023text2room}. We utilize their official codes and pre-trained models for comparison. 
To evaluate the performance of our system, we adopt $21$ scene settings. This includes $6$ challenging outdoor settings ``mountain'', ``garden'', ``house'', ``river'', ``waterfall'', and ``forest'' as well as $15$ indoor settings ``baby room'', ``bathroom'', ``bedroom'', ``cave'', ``forge'', ``ice castle'', ``library'', ``living room'', ``farmhouse living room'', ``modern living room'', ``bedroom-bathroom combo'', ``kitchen-living room combo'', ``large office'', ``small office'', and ``spaceship''. 
Note that we only use LOR~\cite{ren2022look} to generate indoor scenes. For SceneDreamer~\cite{chen2023scenedreamer} and Persistent Nature~\cite{chai2023persistent}, we only utilize them to generate the ``mountain'' scene.
For ours and Text2Room~\cite{hollein2023text2room}, we randomly generate outdoor scenes twice and indoor scenes once, resulting in a total of $12$ outdoor scenes and $15$ indoor scenes. For each scene, we render $200$ images to compute both Inception Score and CLIP Score. For a fair evaluation, when compared with baselines such as Text2Room, we use identical camera trajectories and text prompts as Text2Room to generate 3D scenes and compute quantitative metrics.

\section{Additional Results}
\begin{figure*}[htbp]
    \centering
    \includegraphics[width=\textwidth]{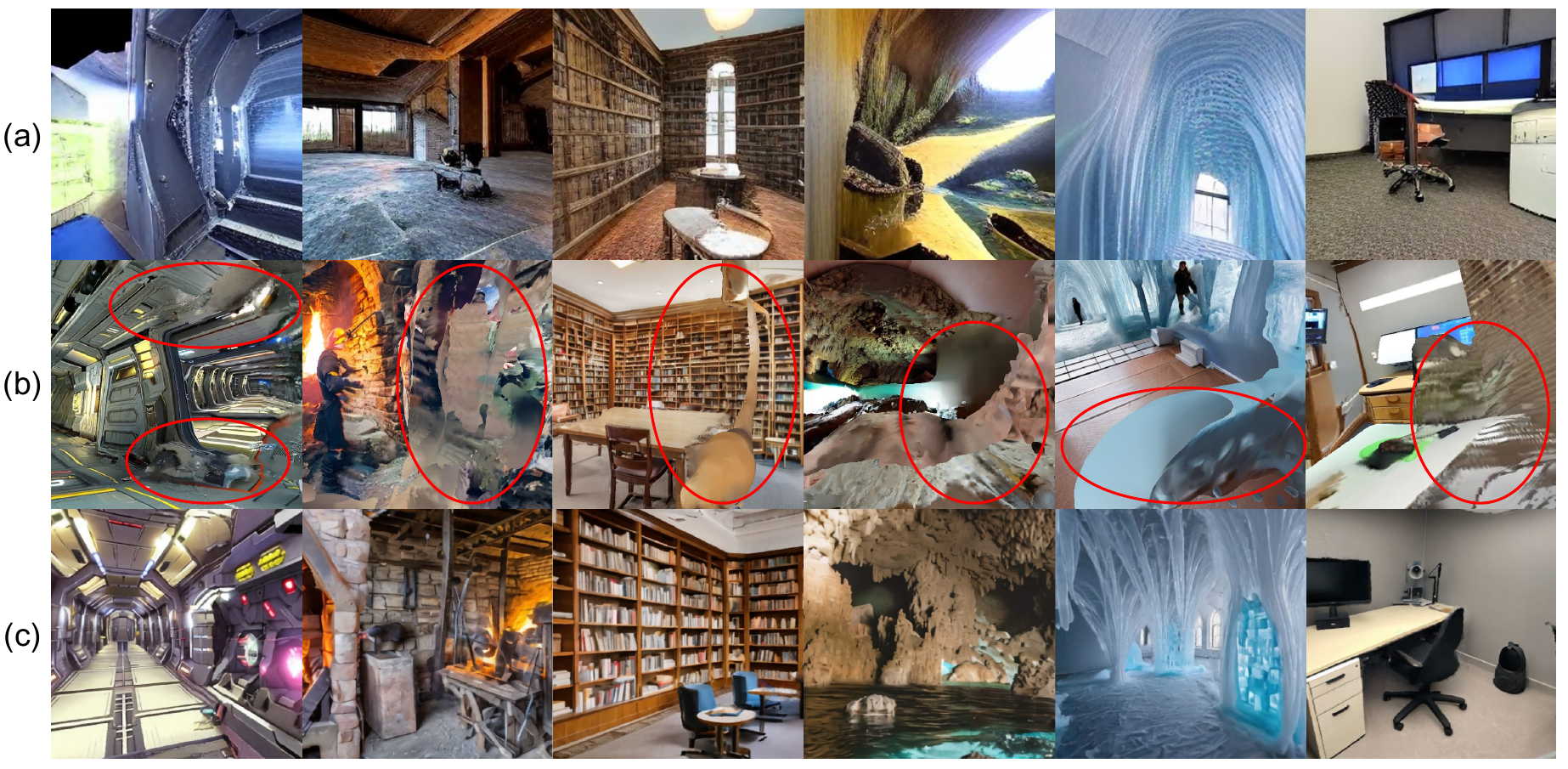}
    \caption{ 
    Additional qualitative comparison on indoor scenes. 
    Here we present the qualitative results of six indoor scenes, displayed alternately from left to right. The scenes, in sequence, are ``spaceship'', ``forge'', ``library'', ``cave'', ``ice castle'', and ``small office''. As can be seen, (a) LOR~\cite{ren2022look} is prone to domain drifting and a decline in output quality. Although (b) Text2Room~\cite{hollein2023text2room} performs well on indoor scenes, it 
    often produces over-smoothed regions and artifacts in the reconstructions. In contrast, (c) our system presents diverse and photo-realistic results.}
    \label{fig:qualitative-indoor-supp}
\end{figure*}

\begin{figure*}[htbp]
    \centering
    \includegraphics[width=\textwidth]{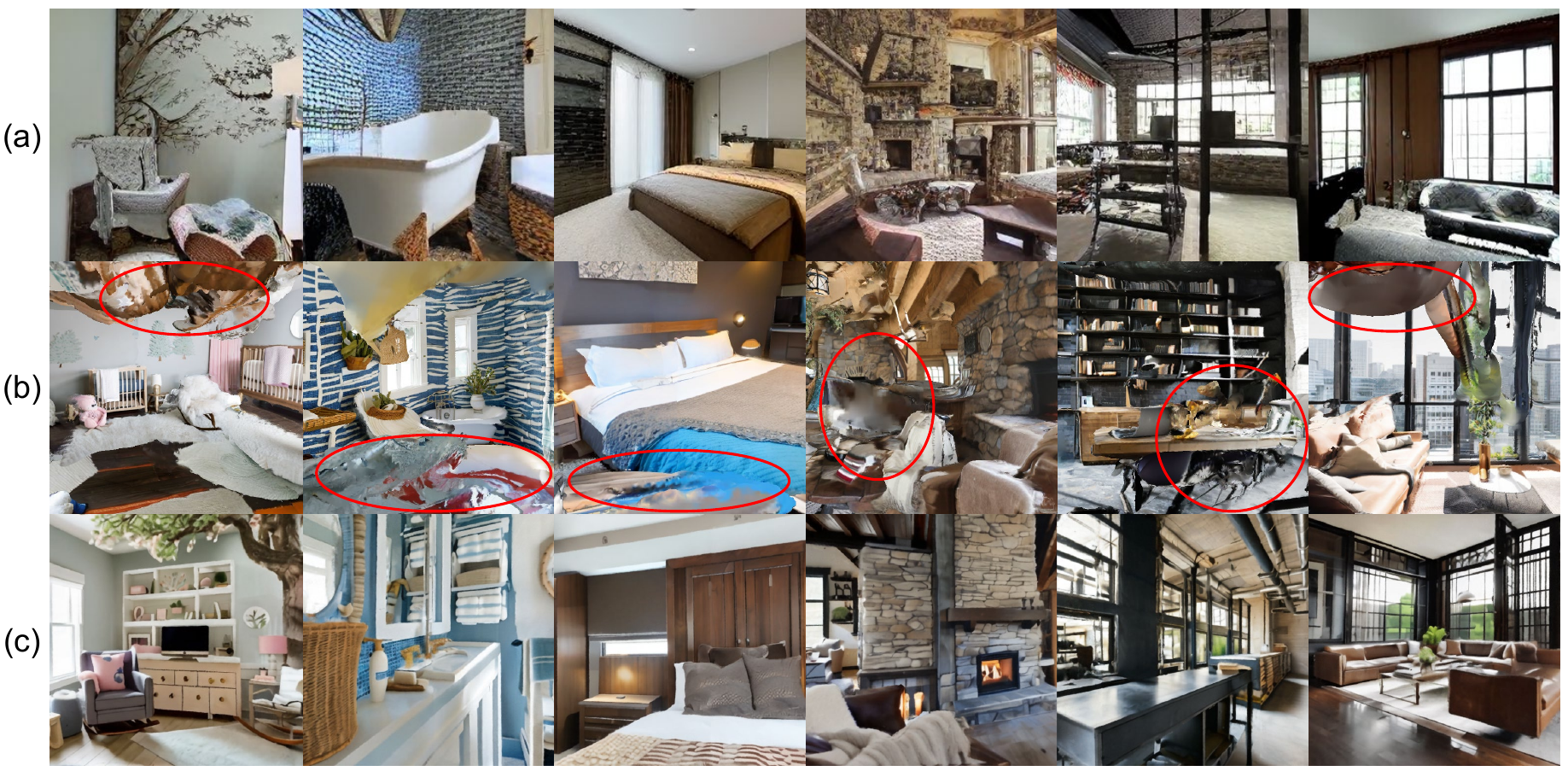}
    \caption{ 
    Additional qualitative comparison on indoor scenes. 
    We present the qualitative results of another six indoor scenes, displayed alternately from left to right. The scenes, in sequence, are ``baby room'', ``bathroom'', ``bedroom'', ``farmhouse living room'', ``large office'', and ``modern living room''. (a) LOR~\cite{ren2022look}, (b) Text2Room~\cite{hollein2023text2room}, and (c) ours.}
    \label{fig:qualitative-indoor-supp2}
\end{figure*}

\begin{figure*}[htbp]
    \centering
    \includegraphics[width=\textwidth]{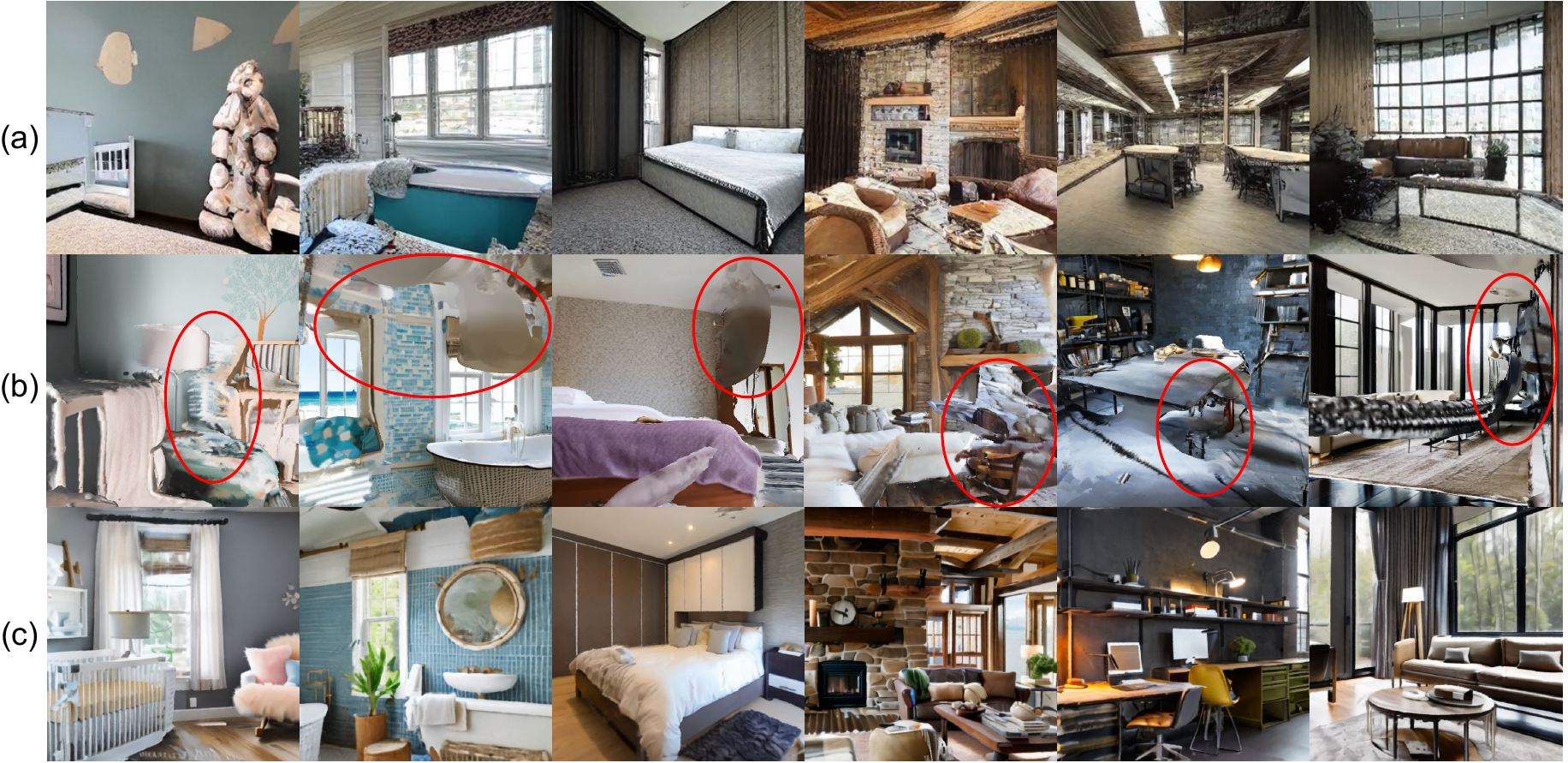}
    \caption{
    Additional qualitative comparison on indoor scenes.  
    We present the qualitative results of another six indoor scenes, displayed alternately from left to right. The scenes, in sequence, are ``baby room'', ``bathroom'', ``bedroom'', ``farmhouse living room'', ``large office'', and ``modern living room''. (a) LOR~\cite{ren2022look}, (b) Text2Room~\cite{hollein2023text2room}, and (c) ours.}
    \label{fig:qualitative-indoor-supp3}
\end{figure*}

\begin{figure*}[htbp]
    \centering
    \includegraphics[width=\textwidth]{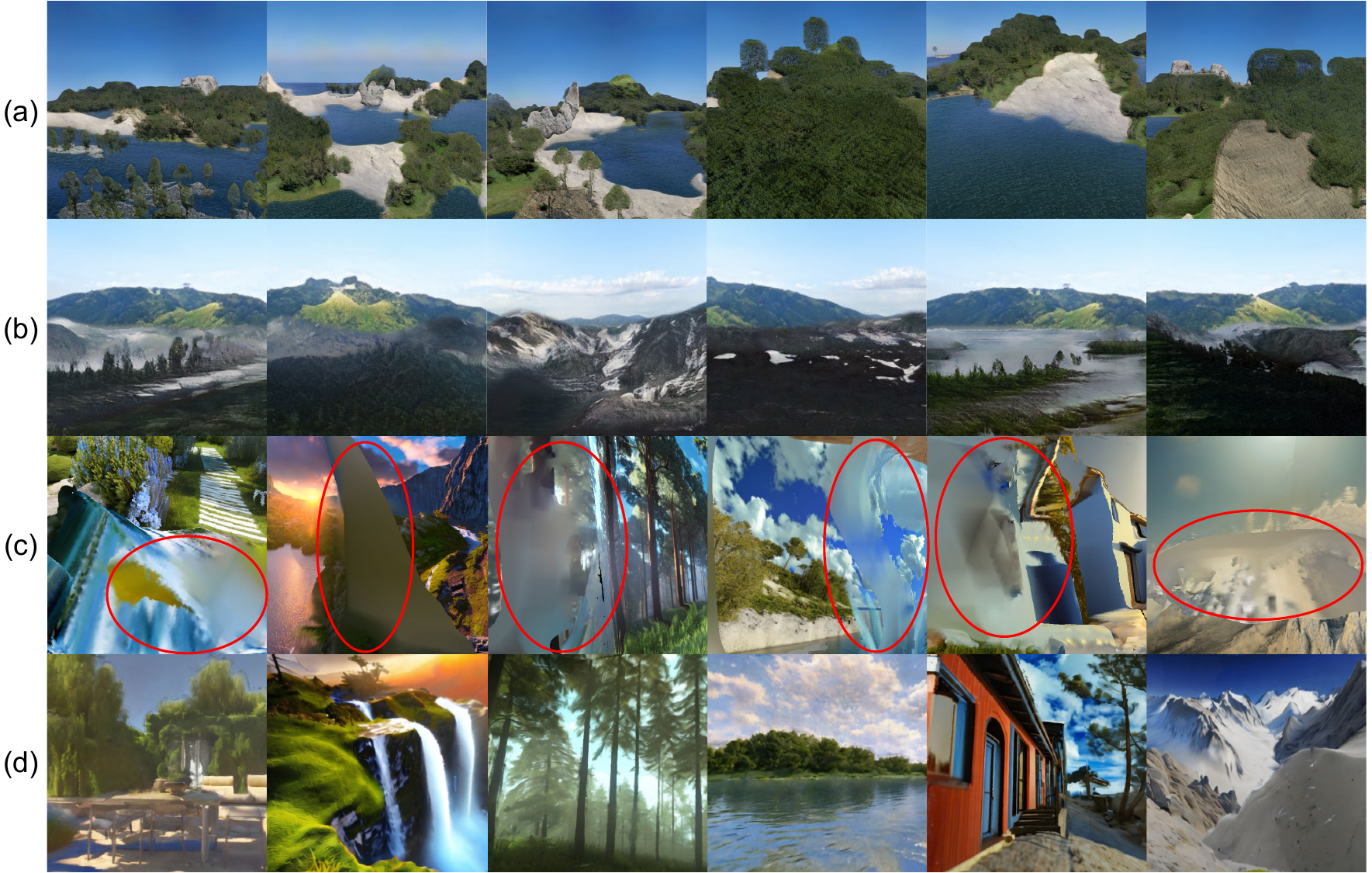}
    \caption{
    Additional qualitative comparison on outdoor scenes. 
    We present the qualitative results of six outdoor scenes, displayed alternately from left to right. The scenes, in sequence, are ``garden'', ``waterfall'', ``forest'', ``river'', ``house'', and ``mountain''. Note that (a) SceneDreamer~\cite{chen2023scenedreamer} and (b) Persistent Nature~\cite{chai2023persistent} require extensive training and are limited to a specific domain, i.e., landscapes. While (c) Text2Room~\cite{hollein2023text2room} can also generate outdoor scenes, it suffers from notable mesh distortions and artifacts. By contrast, (d) our system can generate high-quality and consistent novel views across diverse domains. }
    \label{fig:qualitative-outdoor-supp}
\end{figure*}

\begin{figure*}[htbp]
    \centering
    \includegraphics[width=\textwidth]{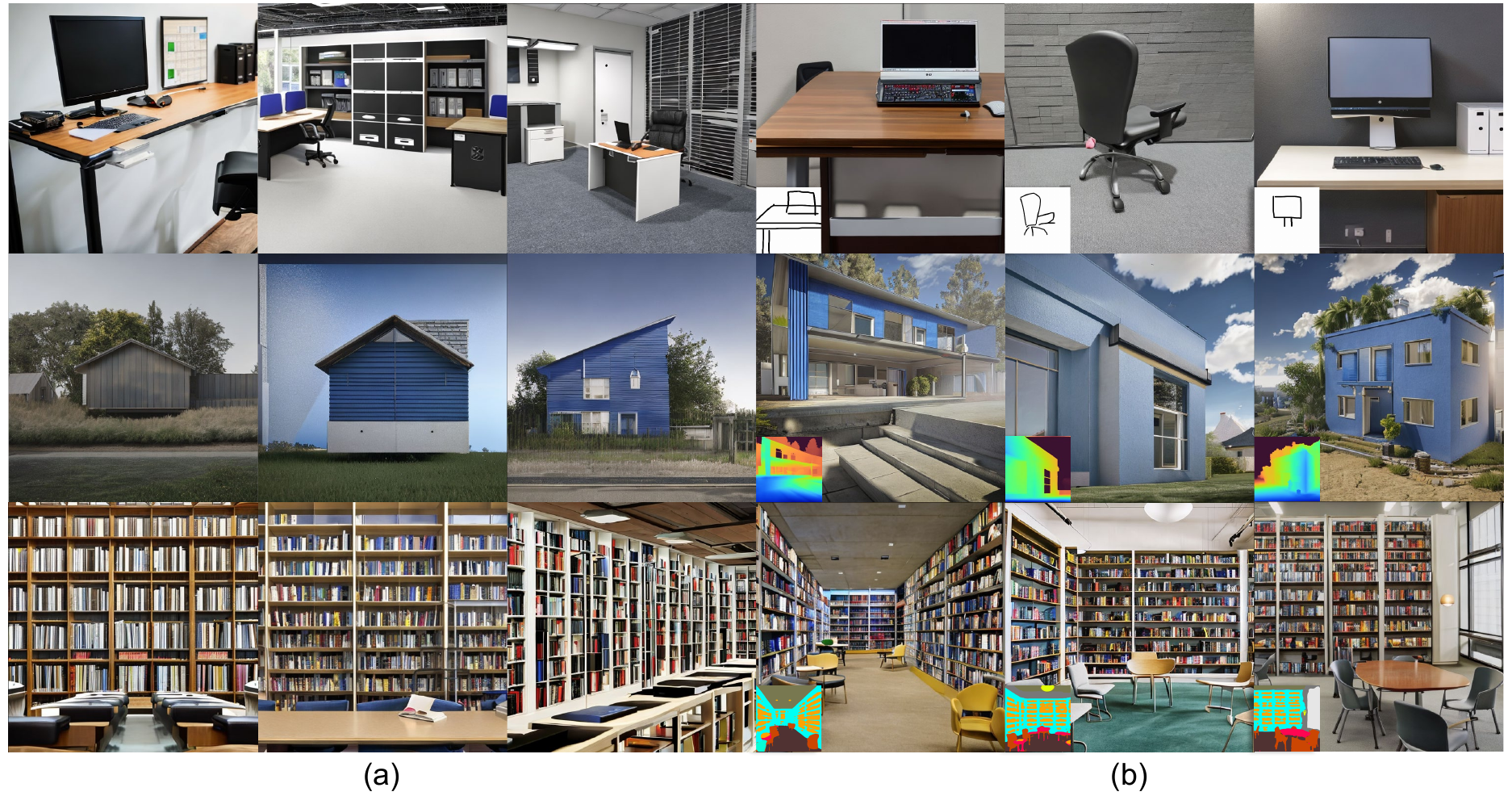}
    \caption{
    Fine-grained control. 
    From top to bottom, we sequentially show ``small office'', ``house'', and ``library''. Compared to (a) current text-driven methods~\cite{hollein2023text2room,fridman2023scenescape}, (b) our system can achieve fine-grained control over the output by adding extra conditions such as scribbles, depth, and semantic segmentation maps. }
    \label{fig:fine-grained-control-supp1}
\end{figure*}

\begin{figure*}[htbp]
    \centering
    \includegraphics[width=\textwidth]{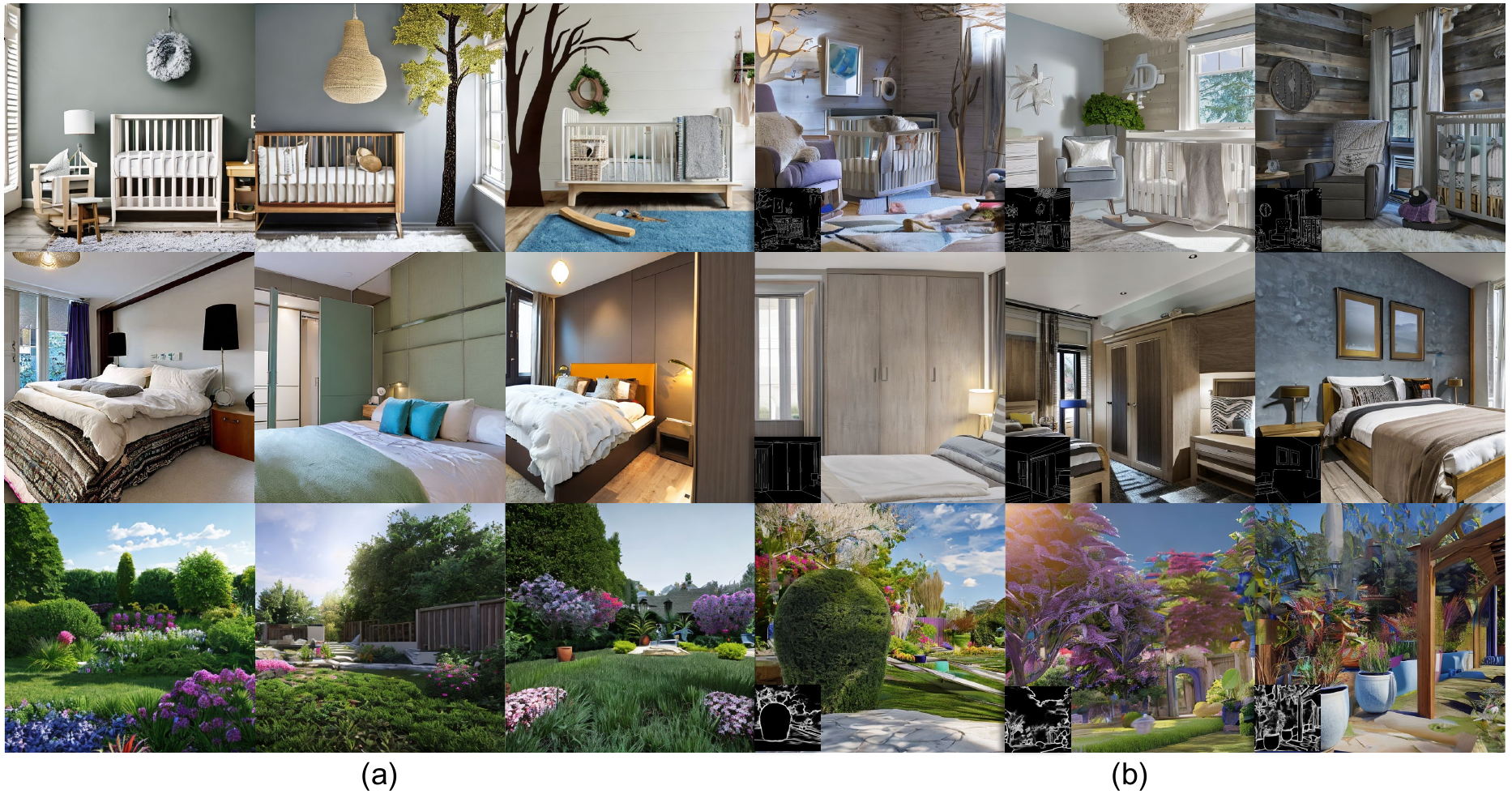}
    \caption{
    Fine-grained control. 
    From top to bottom, we sequentially show ``baby room'', ``bedroom'', and ``garden''. Compared to (a) current text-driven methods~\cite{hollein2023text2room,fridman2023scenescape}, (b) our system can achieve fine-grained control over the output by adding extra conditions such as Canny edge maps, Hough line maps, and HED maps. }
    \label{fig:fine-grained-control-supp2}
\end{figure*}

In this section, we present additional qualitative comparisons in Fig.~\ref{fig:qualitative-indoor-supp}, Fig.~\ref{fig:qualitative-indoor-supp2}, Fig.~\ref{fig:qualitative-indoor-supp3}, and Fig.~\ref{fig:qualitative-outdoor-supp}. As shown in Fig.~\ref{fig:fine-grained-control-supp1} and Fig.~\ref{fig:fine-grained-control-supp2}, besides text prompts, our system allows users to achieve fine-grained control over the output by adding extra conditions such as scribbles, depth maps, semantic segmentation maps, Canny edge maps, Hough line maps, and HED maps.

\section{User Study}
\begin{figure*}[htbp]
    \centering
    \includegraphics[width=\textwidth]{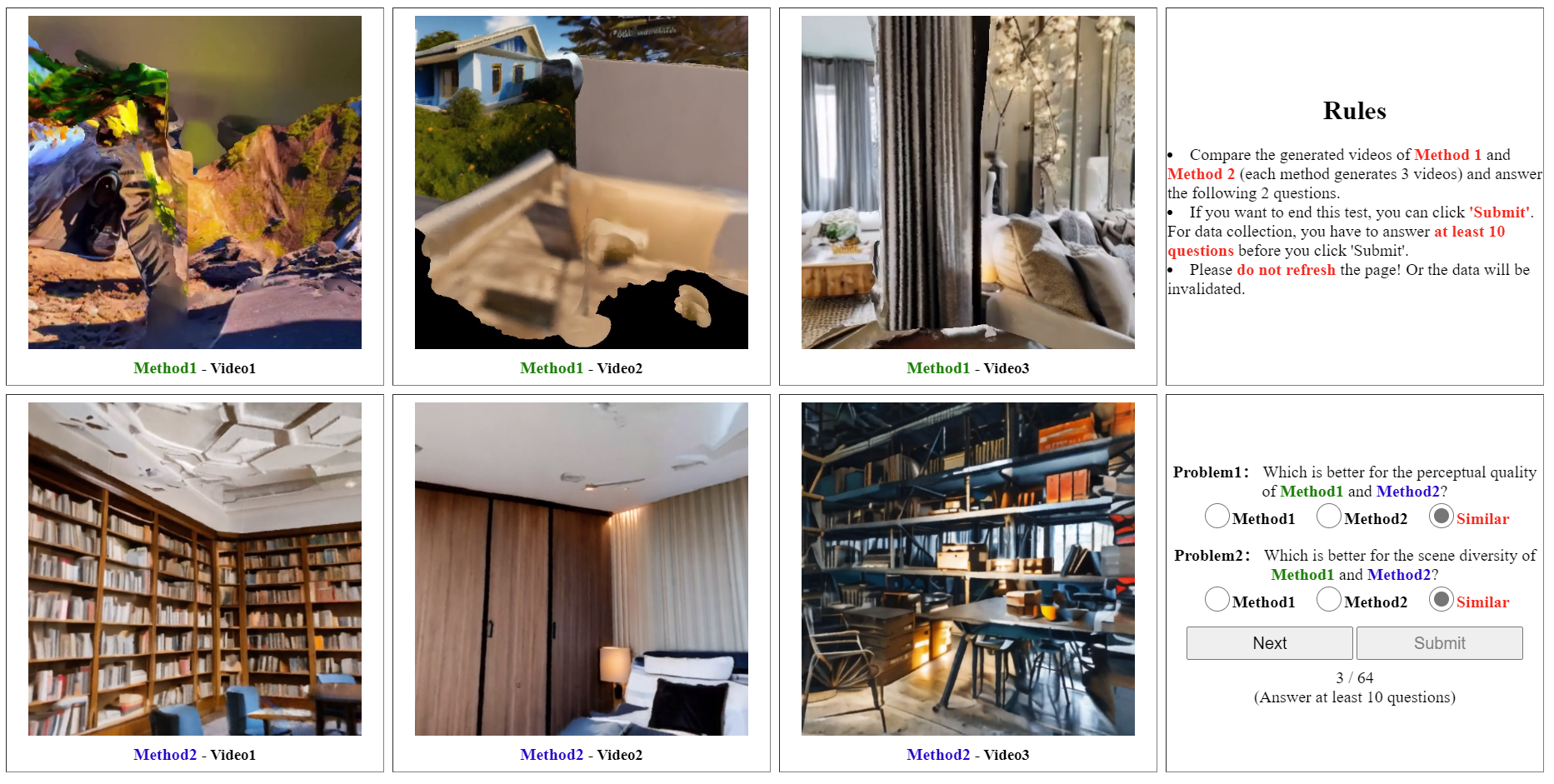}
    \caption{
    Interface of the user study website. 
    }
    \label{fig:user-study-interface}
\end{figure*}

To evaluate the performance of our system, we organize a user study involving $65$ participants with diverse backgrounds and expertise in the field. The study is conducted using an online website designed specifically for this purpose. A screenshot of the website interface is shown in Fig.~\ref{fig:user-study-interface}. Note that our user study is completely anonymous, and no personally identifiable data is collected from the participants. During the study, we present participants with synthesized videos generated by two different methods, labeled as ``Method 1'' and ``Method 2''. To ensure fairness and eliminate bias, each time we randomly select two sets of three videos, where both sets consist of videos generated by randomly chosen methods, including LOR~\cite{ren2022look}, Persistent Nature~\cite{chai2023persistent}, SceneDreamer~\cite{chen2023scenedreamer}, Text2Room~\cite{hollein2023text2room}, and ours, rather than having one set generated exclusively by our system and the other set by another method. This prevents participants from guessing which results are generated by ours during the user study. Participants are asked to compare two key aspects of the videos: the perceptual quality of the imagery and scene diversity. Specifically, they are invited to choose the method that exhibits better perceptual quality and scene diversity or select ``Similar'' if it is difficult to judge. We have a total of $65$ participants, and we collect a substantial amount of data, $2142$ data points in total. On average, each participant answered approximately $33$ questions. In the user study of our main paper, we exclude data points with ``Similar'' options. As a reference, here we provide the version with ``Similar'' options accounted in Table~\ref{tab:user-study}.

\setlength{\tabcolsep}{0\linewidth}
\begin{table}[t]
\footnotesize
    \centering
    \caption{
    User study with ``Similar'' options accounted. 
    All methods are evaluated on the perceptual quality (PQ) of the imagery and scene diversity (SD). Here we only present pairwise comparison results between our system and baselines.}
    \begin{tabular}{C{0.5\linewidth}C{0.25\linewidth}C{0.25\linewidth}}
    \toprule
    Comparison & PQ $\uparrow$  & SD $\uparrow$\\
    \midrule 
    LOR~\cite{ren2022look} / Ours & 15.8\% / \textbf{84.2\%}   & 10.34\% / \textbf{89.7\%}\\ 
    SceneDreamer~\cite{chen2023scenedreamer} / Ours & 36.8\% / \textbf{63.2\%}  & 20.0\% / \textbf{80.0\%}\\ 
    Persistent Nature~\cite{chai2023persistent} / Ours  & 29.8\% / \textbf{70.2\%}  & 21.3\% / \textbf{78.7\%}\\ 
    Text2Room~\cite{hollein2023text2room} / Ours & 27.2\% / \textbf{72.8\%}   & 36.1\% / \textbf{63.9\%} \\ 
    \bottomrule
    \end{tabular}
    \label{tab:user-study}
\end{table}

\section{Why 3D Meshes as an Intermediate Proxy?}
The primary reason for using meshes is that they provide an explicit representation that allows for the iterative build-up of the scene. On the other hand, point clouds are collections of individual points in 3D space, lacking structural information like connectivity and orientation between points. While they are useful for certain tasks like point cloud-based object recognition, they lack the explicit structure necessary for scene creation. Voxel grids divide the 3D space into small cubes or voxels, each representing a discrete volume element. While they offer a more straightforward representation for volumetric data, they often require high memory usage and can be less flexible in handling detailed geometry and shape variations. Implicit representations like NeRFs, are generally not well-suited because the underlying surface geometry is not explicitly represented, which makes it difficult to manipulate and edit the resulting 3D scene representation. 

\section{Limitation Discussion}
While our system provides a user-friendly platform for interactive 3D content creation, certain challenges can impact its performance. (a) The quality of the scene depends on how users create it. Our system offers users the freedom to create 3D scenes according to their will. However, this may be a double-edged sword. For example, if a user only chooses viewpoints in a ``circular rotation'' without changing the camera position, the generated scene might degenerate into a panorama. If a user selects suboptimal viewpoints, the generated scene might contain artifacts or fail to close the loop, i.e., create a complete scene, due to failure cases of depth alignment. In addition, a scene might remain incomplete, e.g., with holes, because parts of the scene are never viewed by the user; (b) The extent to which users can move the camera during the rendering process depends on how users build their scenes. When users create their scenes, the camera's motion can be varied significantly. If users continuously move the camera away from the world origin and progressively build the world, our system can generate videos with substantial camera motion beyond mere circular rotations. In such cases, the rendered videos showcase diverse perspectives and views. However, if users opt to only rotate the camera without changing its position, our system can still generate novel views, but the camera movement will be limited to rotational and slight positional changes. Nevertheless, it is essential to emphasize that our method is not limited to ``circular rotation''. Users have the flexibility to customize their own generation trajectories, enabling a broader range of camera motions. (c) A challenge arises when the depth prediction module produces inaccurate geometry based on the input image, or when the segmentation model fails to predict with precision. These issues can compromise the quality of the generated 3D scenes; (d) Distortions in the 3D meshes may contribute to inaccuracies and inconsistencies, ultimately affecting the overall realism and quality of the output. We leave them for our future work. However, we believe that the proposed system will empower users to unleash their creativity and may open up exciting possibilities for the field of 3D scene generation.

\end{document}